\RequirePackage[fleqn]{amsmath}
\RequirePackage{fix-cm}
\documentclass[smallcondensed,natbib]{svjour3}     
\smartqed  
\usepackage{graphicx}

\usepackage{newtxtext,newtxmath}

\usepackage{url}

\usepackage[fleqn]{amsmath}
\usepackage{amsfonts}
\usepackage{algorithm2e}
\usepackage{array}
\usepackage[caption=false,font=footnotesize]{subfig}
\usepackage{url}
\usepackage{tabularx}
\usepackage{booktabs}
\usepackage{numprint}
\usepackage{multirow}
\usepackage{xcolor}

\newcommand{\bs}{\boldsymbol}


\def\makeheadbox{\relax}

\title{ 
Scalable Bayesian Preference Learning for Crowds
}

\author{Edwin Simpson 
\and Iryna Gurevych
}

\institute{
Edwin Simpson \and Iryna Gurevych \at
Ubiquitous Knowledge Processing Lab, Dept. of Computer Science, Technische Universit\"at Darmstadt.\\
              \email{\{simpson,gurevych\}@ukp.informatik.tu-darmstadt.de}
}

\date{Received: date}

\begin{document}

\titlerunning{Scalable Bayesian Preference Learning}
\authorrunning{Simpson, E and Gurevych, I}


\maketitle

\begin{abstract}
We propose a scalable Bayesian preference learning method 
for jointly predicting the preferences of individuals as well as the consensus of a crowd
 from pairwise labels.
Peoples' opinions often differ greatly,
making it difficult to predict their preferences from small amounts of personal data.
Individual biases also make it harder to infer the consensus of a crowd
when there are few labels per item.
We address these challenges by combining matrix factorisation with 
Gaussian processes,
using a Bayesian approach to account for uncertainty arising from noisy and sparse data.
Our method exploits input features, such as text embeddings and user metadata,
to predict preferences for new items and users that are not in the training set.
As previous solutions based on Gaussian processes do not scale to 
large numbers of users, items or pairwise labels, 
we propose a stochastic variational inference approach that limits computational and memory costs.
Our experiments on a recommendation task show that
our method is competitive with previous approaches despite our scalable inference approximation.
We demonstrate the method's scalability on a natural language processing task 
with thousands of users and items, and show 
improvements over the state of the art on this task.
We make our software publicly available for future 
work~\footnote{\url{https://github.com/UKPLab/tacl2018-preference-convincing/tree/crowdGPPL}}.
\end{abstract}

%


\section{Introduction}\label{sec:intro}

\emph{Preference learning} 
involves comparing a set of alternatives
according to a particular quality~\citep{furnkranz2010preference},
which often leads to a divergence of opinion between people.
For example, in argument mining, 
a sub-field of natural language processing (NLP),
one goal is to rank arguments by their \emph{convincingness}~\citep{habernal2016argument}. 
Whether a particular argument is convincing or not depends on the reader's point of view and prior knowledge~\citep{lukin2017argument}.
Similarly, personal preferences affect recommender systems,
which often perform better if they tailor recommendations
to a specific user~\citep{resnick1997recommender}.
Disagreements also occur when preference annotations are acquired from multiple annotators,
for example, using crowdsourcing,
and are often mitigated by redundant labelling
\citep{snow2008cheap,banerji2010galaxy}.
Therefore, we require preference learning methods that can account for differences of opinion to
(1) predict \emph{personal} preferences for members of a crowd
and
(2) infer a \emph{consensus} given observations from multiple users.
For both tasks, our goal is to rank items or choose the preferred item from any given pair.

Recommender systems often 
predict a user's preferences via \emph{collaborative filtering},  
which overcomes data sparsity by exploiting similarities between the 
preferences of different users~\citep{resnick1997recommender,koren2009matrix}.
Many recommender systems are based on \emph{matrix factorisation} techniques
that are trained using observations of numerical ratings.
However, different annotators often disagree over numerical annotations and can label
inconsistently over time~\citep{ovadia2004ratings,yannakakis2011ranking},
as annotators may interpret the values differently: 
a score of 4/5, say, from one annotator may be equivalent to 3/5 from another. 
The problem is avoided by \emph{pairwise labelling}, 
in which the annotator selects their preferred item from a pair,
which can be quicker~\citep{kendall1948rank,kingsley2010preference,yang2011ranking},
more accurate~\citep{kiritchenko2017best},
and facilitates the total sorting of items, as it avoids two items having the same value.


Pairwise labels provided by a crowd
or extracted from user logs~\citep{joachims2002optimizing} are often noisy and sparse, i.e., many items or users have few or no labels.
This motivates
a Bayesian treatment, 
which has been shown to benefit matrix factorisation
~\citep{salakhutdinov2008bayesian}
and preference learning
~\citep{chen2013pairwise}.
Some previous Bayesian methods for preference learning use
\emph{Gaussian processes (GPs)} to account for \emph{input features} of items or users~\citep{chu2005preference,houlsby2012collaborative,khan2014scalable}.
These are features that can be extracted from content or metadata, such as
\emph{embeddings}~\citep{mikolov2013distributed,devlin2018bert},
which are commonly used by NLP methods to represent words or documents using a numerical vector.
Input features allow the model to extrapolate to new items or users 
and mitigate labelling errors~\citep{felt2016semantic}.
However, previous Bayesian preference learning methods that account for input features using GPs
do not scale to large numbers of items, users, or pairwise labels, 
as their computational and memory requirements grow with the size of the dataset.


In this paper, we propose a scalable Bayesian approach to pairwise preference learning with 
large numbers of users or annotators. 
Our method, \emph{crowdGPPL},
 jointly models personal preferences and the consensus of a crowd through a combination of
matrix factorisation and Gaussian processes. 
We propose a \emph{stochastic variational inference (SVI)} scheme~\citep{hoffman2013stochastic}
that scales to extremely large datasets, as its
memory complexity and the time complexity of each iteration are 
fixed independently of the size of the dataset.
Our new approach opens the door to novel applications involving very large numbers of users, items and pairwise labels,
that would previously have exceeded computational or memory resources and were difficult to parallelise.
We evaluate the method empirically on two real-world datasets to demonstrate 
the scalability of our approach,
and its ability to predict both personal preferences and a consensus given 
preferences from thousands of users.
Our results improve performance over the previous state-of-the-art \citep{simpson2018finding} on a crowdsourced argumentation dataset,
and show that modelling personal preferences improves predictions of the consensus, and vice versa.



\section{Related Work}
\label{sec:rw}

To obtain a ranking from pairwise labels, 
many preference learning methods model
the user's choices as a random function of the latent 
\emph{utility} of the items.
Inferring the utilities of items allows us to rank them, estimate numerical ratings
and predict pairwise labels.
Many popular instances of this approach, known as a \emph{random utility model}~\citep{thurstone1927law},
are variants of  
the Bradley-Terry (BT) model~\citep{bradley1952rank,plackett1975analysis,luce1959possible},
which assumes a logistic likelihood,
or the Thurstone-Mosteller model
~\citep{thurstone1927law,mosteller2006remarks},
which assumes a probit likelihood.
Recent work on the BT model has 
developed computationally efficient active learning, but does not consider input 
features~\citep{li2018hybrid}.
Another commonly-used ranking method, SVM-rank~\citep{joachims2002optimizing},
predicts pairwise labels from input features 
without a random utility model, so cannot predict utilities.
\emph{Gaussian process preference learning (GPPL)}
provides a Bayesian treatment of the random utility model,
using input features to predict the utilities of test items and share information
between similar items~\citep{chu2005preference}.
As GPPL can only predict the preferences of a single user,  
we introduce a new, scalable approach to model individuals in a crowd. 

Previous work on preference learning from crowdsourced data 
treats disagreements as annotation errors and infers only the consensus,
rather than modelling personal preferences.
For instance, 
\citet{chen2013pairwise} and \citet{wang2016blind} 
tackle annotator disagreement 
using Bayesian approaches that learn the labelling accuracy of each worker.
Recently, \citet{pan2018stagewise} and \citet{han2018robust} 
introduced scalable methods that extend this idea from pairwise labels
to noisy \textit{k}-ary preferences,
i.e., totally-ordered subsets of $k$ items.
\citet{fu2016robust} improved SVM-rank by identifying outliers in crowdsourced data
that correspond to probable errors,
while \citet{uchida2017entity} extend SVM-rank to account for different levels of confidence in each pairwise annotation expressed by the annotators.
However, while these approaches differentiate the level of \emph{noise}
for each annotator,
they ignore labelling \emph{bias} 
as the differences between users are not random but depend on personal preferences toward particular items.
With small numbers of labels per item, these biases may reduce the accuracy of the estimated
consensus.
Furthermore, previous aggregation methods for crowdsourced preferences
do not consider item features,
so cannot predict the utility of test items~\citep{chen2013pairwise,wang2016blind,han2018robust,pan2018stagewise,li2018hybrid}.
Our approach goes beyond these methods
by predicting personal preferences
and incorporating input features.

A number of methods use \emph{matrix factorisation} to predict personal preferences 
from pairwise labels, including 
\citet{yi_inferring_2013}, who focus on small numbers of pairs per user,
and \citet{salimans2012collaborative}, who apply Bayesian matrix factorisation to 
handle sparse data.
Matrix factorisation represents observed ratings in a user-item matrix,
which it decomposes into two matrices of lower rank than the user-item matrix, 
one corresponding to users and one to items.
Users with similar ratings have similar columns in the user
matrix, where each entry is a weight over a latent rating.
By multiplying the low-dimensional representations, we can predict ratings for unseen
user-item pairs. 
\citet{kim2014latent} use a simplification that assumes that
each user's preferences depend on only one latent ranking.
However, previous works combining matrix factorisation with
pairwise preference labels do not account for input features.
This contrasts with work on matrix factorisation with side information,
where the ratings or preferences as well as input features are directly observed,
including recent neural network approaches~\citep{NIPS2017_7081},
Bayesian approaches that concatenate input feature vectors with the low-dimensional factored representations~\citep{porteous2010bayesian},
and GP-based methods~\citep{adams2010incorporating}. 
Besides providing a Bayesian method for matrix factorisation with both input features
and pairwise labels, this paper
introduces a much more scalable inference method for a GP-based model.


GPs were previously used 
for personal preference prediction
by \citet{guo2010gaussian}, who propose a GP over the joint feature 
space of users and items. Since this scales cubically
in the number of users, \citet{abbasnejad2013learning} 
propose to cluster users into behavioural groups,
but distinct clusters do not
allow for collaborative learning between users whose preferences only partially overlap, 
e.g. when two users both like one genre of music, 
but have different preferences over other genres. 
\citet{khan2014scalable} instead learn a GP for each user,
then add a matrix factorisation term that performs collaborative filtering.
However, this approach does not model the relationship between
 input features and the low-rank matrices,
 unlike \citet{lawrence2009non} who place GP priors over latent ratings.
 Neither of these last two methods
 are fully Bayesian as the users' weights
 are optimised rather than marginalised.
An alternative is the \emph{collaborative GP (collabGP)}~\citep{houlsby2012collaborative},
which places GP priors over user weights and latent factors,
thereby exploiting input features for both users and items. 
However, unlike our approach, collabGP predicts only pairwise labels, not 
the utilities of items, which are useful for rating and ranking,
and can only be trained using pairwise labels, even if observations of the utilities
are available.
Furthermore, existing GP-based approaches
suffer from scalability issues and 
none of the previous methods jointly model the consensus as well as personal preferences
in a fully-Bayesian manner.

Established methods for GP inference with non-Gaussian likelihoods,
such as the Laplace approximation 
and expectation propagation~\citep{rasmussen_gaussian_2006}, have
time complexity $\mathcal{O}(N^3)$ with $N$ data points
 and memory complexity $\mathcal{O}(N^2)$. 
For collabGP, \citet{houlsby2012collaborative}
use a sparse \emph{generalized fully independent training conditional} (GFITC) 
approximation~\citep{snelson2006sparse} to reduce time complexity to $\mathcal{O}(PM^2 + UM^2)$ and 
memory complexity to $\mathcal{O}(PM + UM)$,
where $P$ is the number of pairwise labels, 
$M \ll P$ is a fixed number of inducing points, 
and $U$ is the number of users.
However, this is not sufficiently scalable
for very large numbers of users or pairs, 
due to increasing memory consumption 
and optimisation steps that cannot be distributed. 
Recent work on distributing and parallelising Bayesian matrix factorisation 
is not easily applicable to models that incorporate GPs
 ~\citep{ahn2015large,saha2015scalable,vander2017distributed,chen2018large}. 
 
To handle large numbers of pairwise labels, \citet{khan2014scalable}
subsample the data rather than learning from the complete training set.
An alternative is \emph{stochastic variational inference (SVI)}~\citep{hoffman2013stochastic}, 
which optimises a posterior approximation using 
a different subsample of training data at each iteration, meaning it learns from
all training data over multiple iterations while limiting costs per iteration.
SVI has been applied to GP regression~\citep{hensman2013gaussian} and classification~\citep{hensman2015scalable},
further improving scalability over earlier sparse approximations.
~\citet{nguyen2014collaborative} introduce SVI for multi-output GPs,
where each output is a weighted combination of latent functions.
They apply their method to capture dependencies between regression tasks,
treating the weights for the latent functions as hyperparameters. 
In this paper, we 
introduce a Bayesian treatment of the weights 
and apply SVI instead to preference learning.
An SVI method for GPPL
was previously introduced by \citet{simpson2018finding},
which we detail in Section \ref{sec:inf}.
However, as GPPL does not consider the individual preferences of users in a crowd,
 we propose a new model, crowdGPPL, which
jointly models personal preferences and the crowd consensus
using a combination of Gaussian processes and Bayesian matrix factorisation.

\section{Bayesian Preference Learning for Crowds}\label{sec:model}

We assume that a pair of items, $a$ and $b$, have utilities
$f(\bs x_a)$ and $f(\bs x_b)$, which represent their value to a user,
and that $f: \mathbb{R}^D \mapsto \mathbb{R}$ 
is a function of item features, where $\bs x_a$ and $\bs x_b$ are vectors 
of length $D$ containing the features of items $a$ and $b$, respectively.
If $f(\bs x_a) > f(\bs x_b)$, then $a$ is preferred to $b$ (written $a \succ b$).
The outcome of a comparison between $a$ and $b$ is 
a pairwise label, $y(a, b)$.
Assuming that pairwise labels never contain errors,
then $y(a, b)=1$ if $a \succ b$ and $0$ otherwise.
Given knowledge of $f$, we can compute the utilities 
of items in a test set given their features, and the outcomes of pairwise comparisons.

\citet{thurstone1927law} proposed the random utility model,
which relaxes the assumption that pairwise labels, $y(a, b)$,
are always consistent with the ordering of $f(\bs x_a)$ and $f(\bs x_b)$.
Under the random utility model, the likelihood $p(y(a,b)=1)$ 
increases as $f_a - f_b$ increases, i.e.,
as the utility of item $a$ increases
relative to the utility of item $b$.
This reflects the greater consistency in a user's choices
when their preferences are stronger,
while accommodating
labelling errors or variations in a user's choices over time.
In the Thurstone-Mosteller model, 
noise in the observations is explained by a Gaussian-distributed noise term, $\delta \sim \mathcal{N}(0, \sigma^2)$:
\begin{flalign}
 p(y(a, b) | f(\bs x_a) + \delta_{a}, f(\bs x_b) + \delta_{b} )  
 \hspace{0.9cm} & = \begin{cases}
 1 & \text{if }f(\bs x_a) + \delta_{a} \geq f(b) + \delta_{b} \\
 0 & \text{otherwise,}
 \end{cases} &
 \label{eq:thurstone}
\end{flalign}
Integrating out the unknown values of $\delta_a$ and $\delta_b$ gives:
\begin{flalign}
& p( y(a, b) | f(\bs x_a), f(\bs x_b) )  & \label{eq:plphi}\\
& = \!\! \int\!\!\!\! \int \!\! p( y(a, b) | f(\bs x_a) + \delta_{a}, f(\bs x_b) + \delta_{b} ) \mathcal{N}\left(\delta_{a}; 0, \sigma^2\right)\mathcal{N}\left(\delta_{b}; 0, \sigma^2\right) d\delta_{a} d\delta_{b} 
= \Phi\left( z \right), & \nonumber
\end{flalign}
where $z = \frac{f(\bs x_a) - f(\bs x_b)}{\sqrt{2\sigma^2}}$,
and $\Phi$ is the cumulative distribution function of the standard normal distribution,
meaning that $\Phi(z)$ is a 
probit likelihood.\footnote{Please note that a full list of symbols is provided for reference in Appendix $\ref{sec:not}$}
This likelihood is also used by
\citet{chu2005preference} for Gaussian process preference learning (GPPL), but here 
we simplify the formulation by assuming that $\sigma^2 = 0.5$,
which leads to $z$ having a denominator of $\sqrt{2 \times 0.5}=1$,
hence $z = f(\bs x_a) - f(\bs x_b)$.
Instead, we model varying degrees of noise in the pairwise labels
by scaling $f$ itself, as we describe in the next section.

In practice, $f(\bs x_a)$ and $f(\bs x_b)$ must be inferred from
pairwise training labels, $\bs y$,
to obtain a posterior distribution over their values.
If this posterior is a multivariate Gaussian distribution,
then the probit likelihood allows us to analytically marginalise 
$f(\bs x_a)$ and $f(\bs x_b)$
to obtain the probability of a pairwise label:
\begin{flalign}
p(y(a,b)| \bs y) 
= \Phi(\hat{z}),& & \hat{z} = \frac{\hat{f}_a - \hat{f}_b}{\sqrt{1 + C_{a,a} + C_{b,b} 
- 2C_{a,b}} }, \label{eq:predict_z} &&
\end{flalign}
where $\hat{f}_a$ and $\hat{f}_b$ are the means and
$\bs C$ is the posterior covariance matrix of the multivariate Gaussian over
$f(\bs x_a)$ and $f(\bs x_b)$.
Unlike other choices for the likelihood, such as a sigmoid,
the probit allows us to compute the posterior over a pairwise label
without further approximation,
hence we assume this pairwise label likelihood for our proposed preference learning model.

\subsection{GPPL for Single User Preference Learning}

We can model the preferences of a single user by assuming
a Gaussian process prior over the user's utility function, 
$f \sim \mathcal{GP}(0, k_{\theta}/s)$, where $k_{\theta}$ is a kernel function with hyperparameters $\theta$
and $s$ is an inverse scale parameter.
The kernel function takes numerical item features as inputs and determines the covariance between values of $f$ for different items. 
The choice of kernel function and its hyperparameters controls the shape and smoothness of the function 
across the feature space and is often treated as a model selection problem.
Kernel functions suitable for a wide range of tasks include the \emph{squared exponential} 
and the \emph{Mat\'ern}~\citep{rasmussen_gaussian_2006},
which both make minimal assumptions but 
assign higher covariance to items with similar feature values.
We use $k_{\theta}$ to compute a covariance matrix $\bs K_{\theta}$,
between a set of $N$ observed items with features $\bs X = \{ \bs x_1, ..., \bs x_N \}$.

Here we extend the original definition of GPPL~\citep{chu2005preference},
by introducing the inverse scale, $s$,
which is drawn from a gamma prior, 
$s \sim \mathcal{G}(\alpha_0, \beta_0)$, with shape $\alpha_0$ and scale $\beta_0$.
The value of $1/s$ determines the variance of $f$,
and therefore 
the magnitude of differences between $f(\bs x_a)$ and $f(\bs x_b)$ for
items $a$ and $b$. This in turn affects the level of certainty
in the pairwise label likelihood as per Equation \ref{eq:plphi}.

Given a set of $P$ pairwise labels, 
$\bs y=\{y_1,...,y_P\}$,
where 
$y_p=y(a_p, b_p)$ is the preference label for items $a_p$ and $b_p$, 
we can write the joint distribution over all variables as follows:
\begin{flalign}
p\left( \bs{y}, \bs f, s | k_{\theta}, \bs X, \alpha_0, \beta_0 \right) 
=  \prod_{p=1}^P p( y_p | \bs f ) 
\mathcal{N}(\bs f; \bs 0, \bs K_{\theta}/s) \mathcal{G}(s; \alpha_0, \beta_0) 
\label{eq:joint_single}
\end{flalign}
where 
$\bs f = \{f(\bs {x}_1),...,f(\bs {x}_N)\}$
is a vector containing the utilities of the $N$ items referred to by $\bs y$,
and $p( y_p | \bs f ) = \Phi\left( z_p \right)$ is the pairwise likelihood (Equation \ref{eq:plphi}). 

\subsection{Crowd Preference Learning} \label{sec:crowd_model}

To predict the preferences of individuals in a crowd,
we could use an independent GPPL model for each user.
However, by modelling all users jointly, we can
exploit correlations between their interests
to improve predictions when preference data is sparse,
and reduce the memory cost of storing separate models.
Correlations between users 
can arise from common interests over certain subsets of items,
such as in one particular genre in a book recommendation task.
Identifying such correlations helps to predict 
 preferences from  fewer observations and is the core idea of collaborative filtering~\citep{resnick1997recommender} and matrix factorisation~\citep{koren2009matrix}.

As well as individual preferences, 
we wish to predict the consensus by aggregating
preference labels from multiple users. 
Individual biases of different users may affect consensus predictions,
particularly when data for certain items comes from a small subset of users.
The consensus could also help
predict preferences of users with little or no data
 by favouring popular items
and avoiding generally poor items.
We therefore propose 
 \emph{crowdGPPL}, which jointly models 
the preferences of individual users as well as the underlying consensus of the crowd.
Unlike previous methods for inferring the consensus, 
such as \emph{CrowdBT}~\citep{chen2013pairwise}, we do not treat differences between users as simply the result of labelling errors, 
but also account for their subjective biases
towards particular items. 
 
For crowdGPPL, 
we represent utilities in a matrix, $\bs{F} \in \mathbb{R}^{N \times U}$,
with 
$U$ columns corresponding to users. 
Within $\bs F$, each entry $F_{a,j} = f(\bs x_a, \bs u_j)$ is the 
utility for item $a$ for user $j$ with user features $\bs u_j$.
We assume that $\bs{F} = \bs{V}^T \bs{W} + \bs{t}\bs{1^T}$
 is the product of two low-rank matrices
plus a column vector of consensus utilities, $\bs{t} \in \mathbb{R}^N$, 
where $\bs{W} \in \mathbb{R}^{C \times U}$ is a latent representation
of the users,
$\bs{V} \in \mathbb{R}^{C \times N}$ is a latent representation of the items,
 $C$ is the number of latent \emph{components}, i.e., the dimension
of the latent representations,
and $\bs 1$ is a column vector of ones of length $U$. 
The column $\bs v_{.,a}$ of $\bs V$, and the column $\bs w_{.,j}$ of $\bs W$,
 are latent vector representations of item $a$ and user $j$,
 respectively.
Each row of $\bs V$, $\bs v_c=\{ 
v_c(\bs{x}_1),...,v_c(\bs{x}_N)\}$,  
contains evaluations of a latent function, 
$v_c\sim \mathcal{GP}(\bs 0, k_{\theta} /s^{(v)}_c)$,
of item features, $\bs x_a$,
where $k$ is a kernel function, $s^{(v)}_c$ is an inverse function scale,
and $\theta$ are kernel hyperparameters.
The consensus utilities, $\bs t = \{t(\bs {x}_1),...,t(\bs {x}_N)\}$,
are values of a consensus utility function over item features,
$t\sim \mathcal{GP}(\bs 0, k_{\theta} /s^{(t)})$, which is shared across all users,
with inverse scale $s^{(t)}$.
Similarly, each row of $\bs W$, 
$\bs w_c=\{w_c(\bs u_1),...,w_c(\bs u_U)\}$,
 contains evaluations of a latent function,
$w_c \sim \mathcal{GP}(\bs 0, k_{\eta}/s_c^{(w)})$,
of user features, $\bs u_j$, 
with inverse scale $s_c^{(w)}$
and kernel hyperparameters $\eta$.
Therefore, each utility in $\bs F$ can be written as
a weighted sum over the latent components:
\begin{flalign}
  f(\bs x_a, \bs u_j) = \sum_{c=1}^C  v_c(\bs x_a) w_c(\bs u_j) + t(\bs x_a),
  \label{eq:vw_plus_t}
\end{flalign}
where $\bs u_j$ are the features of user $j$ and $\bs x_a$ are the features of item $a$.
Each latent component corresponds to a utility function 
for certain items, which is shared by a subset of users to differing degrees.
For example, in the case of book recommendation,
$c$ could relate to science fiction novels, 
$v_c$ to a ranking over them,
and $w_c$ to the degree of agreement of users with that ranking.
The individual preferences of each user $j$ deviate from a consensus across users, $t$, according
to $\sum_{c=1}^C  v_c(\bs x_a) w_c(\bs u_j)$. 
This allows us to subtract the effect of individual biases when inferring the consensus utilities. 
The consensus can also help 
when inferring personal preferences for 
new combinations of users and items that are
very different to those in the training data by
 accounting for any objective or widespread appeal that an item may have.

Although the model assumes a fixed number of components, $C$,
the GP priors over $\bs w_c$ and $\bs v_c$ act as \emph{shrinkage}
or \emph{ARD priors} that favour values close to zero~\citep{mackay1995probable,psorakis2011overlapping}. 
Components that are not required to explain the data will have posterior
expectations and scales $1/s^{(v)}$ and $1/s^{(w)}$ approaching zero.
Therefore, 
it is not necessary to optimise the value of $C$ by hand, 
providing a sufficiently large number is chosen. 

Equation \ref{eq:vw_plus_t} is similar to
\emph{cross-task crowdsourcing}~\citep{mo2013cross}, which 
uses matrix factorisation to model annotator performance in different tasks,
where $\bs t$ corresponds to the objective difficulty of a task.
However, unlike crowdGPPL, they do not use GPs to model the factors, 
nor apply
the approach to preference learning.
For preference learning, collabGP~\citep{houlsby2012collaborative}
is a related model that 
excludes the consensus and uses values in $\bs v_c$ to represent pairs
 rather than individual items, so does not infer item ratings.
It also omits scale parameters for the GPs that 
encourage shrinkage when $C$ is larger than required.
 
We combine the matrix factorisation method with the preference likelihood of Equation \ref{eq:plphi}
to obtain the joint preference model for multiple users, \emph{crowdGPPL}:
\begin{flalign}
&p\left( \bs{y}, \bs V, \bs W, \bs t, s^{(v)}_1 \!\!, .., s^{(v)}_C\!\!, s^{(w)}_1\!\!, .., s^{(w)}_C\!\!, s^{(t)} 
| k_{\theta}, \bs X, k_{\eta}, \bs U, \alpha_0^{(t)}\!\!, \beta_0^{(t)}\!\!,
\alpha_0^{(v)}\!\!, \beta_0^{(v)}\!\!, \alpha_0^{(w)}\!\!, \beta_0^{(w)} \right) 
 & \nonumber \\ 
& = \prod_{p=1}^P \Phi\left( z_p \right) 
\mathcal{N}\left(\bs t; \bs 0, \bs K_{\theta} /s^{(t)}\right)
\mathcal{G}\left({s^{(t)}}; \alpha_0^{(t)}, \beta_0^{(t)}\right)
\prod_{c=1}^C \left\{
\mathcal{N}\left(\bs v_c; \bs 0, \bs K_{\theta} /s^{(v)}_c\right)
\right.
 & \nonumber \\  
&\left.
\mathcal{N}\left(\bs w_c; \bs 0, \bs L_{\eta}/s^{(w)}_c\right) \mathcal{G}\left(s^{(v)}_c; \alpha_0^{(v)}, \beta_0^{(v)}\right)\mathcal{G}\left(s^{(w)}_c; \alpha_0^{(w)}, \beta_0^{(w)} \right) \right\}, &
\label{eq:joint_crowd}
\end{flalign}
where 
$z_p = \bs v_{.,a_p}^T \bs{w}_{.,u_p} + t_{a_p} - \bs v_{.,b_p}^T \bs{w}_{.,u_p} - t_{b_p}$,
index $p$ refers to a user and a pair of items, $\{u_p, a_p, b_p \}$,
$\bs U$ is the set of feature vectors for all users,
$\bs K_{\theta}$ is the prior covariance for the items as in GPPL,
and $\bs L_{\eta}$ is the prior covariance for the users computed
using $k_{\eta}$.

\section{Scalable Inference}\label{sec:inf}

Given a set of pairwise training labels, $\bs y$,
we aim to find the posterior over the matrix
$\bs F^*=\bs V^{*T} \bs W^*$ of utilities for test items and test users,
and the posterior over consensus utilities for test items, $\bs t^*$.
The non-Gaussian likelihood (Equation \ref{eq:plphi})
makes exact inference intractable, hence previous work uses
 the Laplace approximation for GPPL~\citep{chu2005preference}
or combines expectation propagation (EP) with variational Bayes for a 
multi-user model~\citep{houlsby2012collaborative}.
The Laplace approximation is a maximum a-posteriori solution that
takes the most probable values of parameters rather than integrating over their distributions,
and has been shown to perform poorly for classification compared to EP~\citep{nickisch2008approximations}. 
However, 
a drawback of EP is that convergence is not guaranteed
~\citep{minka2001expectation}.
More importantly, inference for a GP using either method
has computational complexity $\mathcal{O}(N^3)$ 
and memory complexity $\mathcal{O}(N^2)$, where $N$ is the number of data points.

The cost of inference can be reduced using a \emph{sparse} approximation based on a set of 
\emph{inducing points}, which act as substitutes for the points in the training dataset.
By choosing a fixed number of inducing points, $M \ll N$, the computational cost is cut to $\mathcal{O}(NM^2)$,
and the memory complexity to $\mathcal{O}(NM)$.
Inducing points must be selected 
using either heuristics or by optimising their positions to maximise an estimate of the 
marginal likelihood. 
One such sparse approximation is the \emph{generalized fully independent training conditional} (GFITC)~\citep{NIPS2007_3351,snelson2006sparse}, 
used by \citet{houlsby2012collaborative} for collabGP.
However, time and memory costs that grow linearly with $\mathcal{O}(N)$
start to become a problem with thousands of data points,
as all data must be processed in every iterative update,
before any other parameters such as $s$ are updated,
making GFITC unsuitable for very large datasets~\citep{hensman2015scalable}.

We derive a more scalable approach for GPPL and crowdGPPL using
stochastic variational inference (SVI)~\citep{hoffman2013stochastic}.
For GPPL, this reduces the time complexity of each iteration 
to $\mathcal{O}(P_i M^2 + P_i^2 M + M^3)$,
and memory complexity 
to $\mathcal{O}(P_i M + M^2  + P_i^2)$,
where $P_i$ is a mini-batch size that we choose in advance.
Neither $P_i$ nor $M$ are dependent on the size of the dataset, meaning that SVI 
can be run with arbitrarily large datasets, 
and other model parameters such as $s$ can be updated before processing all data
to encourage faster convergence.
First, we define a suitable likelihood approximation to enable the use of SVI.

\subsection{Approximating the Posterior with a Pairwise Likelihood}

The preference likelihood in Equation \ref{eq:plphi} 
is not conjugate with the Gaussian process, which means there is no analytic expression for
the exact posterior.
For single-user GPPL, we therefore
approximate the preference likelihood with a Gaussian:
\begin{flalign}
p(\bs f | \bs y, s) & \propto \prod_{p=1}^P p\left(y_p | z_p\right) p\left(\bs f | \bs K, s\right)
= \prod_{p=1}^P \Phi\left(z_p\right) \mathcal{N}\left(\bs f; \bs 0, \bs K/s\right)
& \\
& \approx \prod_{p=1}^P \mathcal{N}\left(y_p; \Phi(z_p), Q_{p,p}\right) 
\mathcal{N}\left(\bs f; \bs 0, \bs K/s\right)
 = \mathcal{N}\left(\bs y; \Phi(\bs z), \bs Q\right) \mathcal{N}\left(\bs f; \bs 0, \bs K/s\right), &\nonumber 
\end{flalign}
where $\bs Q$ is a diagonal noise covariance matrix
and we omit the kernel hyperparameters, $\theta$, to simplify notation.
For crowdGPPL, we use the same approximation to the likelihood, but
replace $\bs f$ with $\bs F$.
We estimate the diagonals of $\bs Q$ 
by moment matching our approximate likelihood with $\Phi(z_p)$,
which defines a Bernoulli distribution with variance $Q_{p,p} = \Phi(z_p)(1 - \Phi(z_p))$.
However, this means that $\bs Q$ 
depends on $\bs z$ and therefore on $\bs f$,
so the approximate posterior over $\bs f$ cannot be computed in closed form.
To resolve this, we approximate $Q_{p,p}$ 
using an estimated posterior over $\Phi(z_p)$ computed
independently for each pairwise label, $p$.
We obtain this estimate
 by updating the parameters of the conjugate prior for the Bernoulli likelihood,
 which is
a beta distribution with parameters $\gamma$ and $\lambda$.
We find $\gamma$ and $\lambda$ by 
matching the moments of the beta prior to the prior mean and variance of $\Phi(z_p)$,
estimated using numerical integration.
The prior over $\Phi(z_p)$ is defined by a GP for single-user GPPL, $p(\Phi(z_p) | \bs K, \alpha_0, \beta_0)$,
and a non-standard distribution for crowdGPPL. 
Given the observed label $y_p$, we estimate the diagonals in $\bs Q$
as the variance of the posterior beta-Bernoulli:
\begin{flalign}
Q_{p,p} & \approx \frac{ (\gamma + y_p)(\lambda + 1 - y_p) }{(\gamma + \lambda + 1)^2}. &
\end{flalign}
The covariance $\bs Q$ therefore approximates the expected noise in the observations, 
hence captures variance due to $\sigma$ in Equation \ref{eq:plphi}.
This approximation performs well empirically
for Gaussian process classification~\citep{reece2011determining,simpson2017bayesian} and 
classification using extended Kalman filters~\citep{lee2010sequential,lowne2010sequential}. 

Unfortunately, the nonlinear term $\Phi(\bs z)$ means that the posterior is still intractable, 
so we replace $\Phi(\bs z)$ with a linear function of $\bs f$ by taking
the first-order Taylor series expansion of $\Phi(\bs z)$ 
about the expectation $\mathbb{E}[\bs f] = \hat{\bs f}$:
\begin{flalign}
\Phi(\bs z) &\approx \tilde{\Phi}(\bs z) = \bs G \left(\bs f-\hat{\bs f}\right) 
+ \Phi(\hat{\bs z}), & \\
G_{p,i} &= \frac{\partial \Phi(\hat{z}_p)} {\partial f_i}
= \Phi(\hat{z}_p)\left(1 - \Phi(\hat{z}_p)\right) \left(2y_p - 1\right)\left( [i = a_p] - [i = b_p]\right), &
\end{flalign}
where $\hat{\bs z}$ is the expectation of $\bs z$ computed using Equation \ref{eq:predict_z},
and $[i=a]=1$ if $i=a$ and is $0$ otherwise. 
There is a circular dependency between $\hat{\bs f}$,
which is needed to compute $\hat{\bs z}$, and $\bs G$. 
We estimate these terms using a variational inference procedure
that iterates between updating $\bs f$ and $\bs G$~\citep{steinberg2014extended}
as part of Algorithm \ref{al:singleuser}.
The complete approximate posterior for GPPL is now as follows:
\begin{flalign}
p(\bs f | \bs y, s) 
\approx 
\mathcal{N}\left(\bs y; \bs G (\bs f-\mathbb{E}[\bs f]) + \Phi(\hat{\bs z}), \bs Q\right) \mathcal{N}\left(\bs f; \bs 0, \bs K/s\right) / Z = \mathcal{N}\left(\bs f; \hat{\bs f}, \bs C\right), &&
\label{eq:likelihood_approx} 
\end{flalign}
where $Z$ is a normalisation constant.
Linearisation means that our approximate likelihood is conjugate to the prior,
so the approximate posterior is also Gaussian. 
Gaussian approximations to the posterior have shown strong empirical results for 
classification~\citep{nickisch2008approximations} and
preference learning~\citep{houlsby2012collaborative},
and linearisation using a Taylor expansion has been widely tested
in the extended Kalman filter~\citep{haykin2001kalman}
as well as Gaussian processes~\citep{steinberg2014extended,bonilla2016extended}.

\subsection{SVI for Single User GPPL}

Using the linear approximation in the previous section, 
posterior inference requires inverting
$\bs K$ with computational cost $\mathcal{O}(N^3)$
and taking an expectation with respect to $s$, which remains intractable. 
We address these problems using stochastic variational inference (SVI)
with a sparse approximation to the GP that limits
the size of the covariance matrices we need to invert.
We introduce $M \ll N$ inducing items with inputs 
$\bs X_m$,
utilities $\bs f_m$, and covariance $\bs K_{mm}$. The
covariance between the observed and inducing items is $\bs K_{nm}$.
For clarity, we omit $\theta$ from this point on.
We assume a \emph{mean-field} approximation to the joint posterior over 
inducing and training items
that factorises between different sets of latent variables:
\begin{flalign}
p\left(\bs f, \bs f_m, s | \bs y, \bs X, \bs X_m, k_{\theta}, \alpha_0, \beta_0 \right) 
&\approx q\left(\bs f, \bs f_m, s\right) = q(s)q\left(\bs f\right)q\left(\bs f_m\right), \label{eq:svi_approx} &&
\end{flalign}
where $q(.)$ are \emph{variational factors} defined below. 
Each factor corresponds to a subset of latent variables, $\bs \zeta_i$, and
takes the form $\ln q(\bs \zeta_i) = \mathbb{E}_{j \neq i}[\ln p(\bs \zeta_i, \bs x, \bs y)]$.
That is, the expectation with respect
to all other latent variables, $\bs\zeta_j,\forall j \neq i$, of the log joint distribution
of the observations and latent variables, $\bs \zeta_i$.
To obtain the factor for $\bs f_m$, we marginalise $\bs f$ and take expectations with respect to $q(s)$:
\begin{flalign}
\ln q\left(\bs f_m\right) &= \ln \mathcal{N}\!\left(\bs y; \tilde{\Phi}(\bs z), \bs Q\right)
+ \ln\mathcal{N}\left(\bs f_m; \bs 0, \frac{\bs K_{mm}}{\mathbb{E}\left[s\right]}\right) \!  + \textrm{const} 
 = \ln \mathcal{N}\left(\bs f_m; \hat{\bs f}_m, \bs S \right), &
 \label{eq:fhat_m}
\end{flalign}
where the variational parameters $\hat{\bs f}_m$ and $\bs S$ are computed using 
an iterative SVI procedure described below.
We choose an approximation of $q(\bs f)$ that depends only on the inducing point utilities, $\bs f_m$, and is independent of the observations:
 \begin{flalign}
\ln q\left(\bs f\right) & = \ln \mathcal{N}\left(\bs f; \bs A \hat{\bs f}_m, 
\bs K + \bs A \left(\bs S - \bs K_{mm}/\mathbb{E}[s]\right) \bs A^T \right), &
\end{flalign}
where $\bs A=\bs K_{nm} \bs K^{-1}_{mm}$.
Therefore, we no longer need to invert an $N \times N$ covariance matrix to compute $q(\bs f)$.
The factor $q(s)$ also depends only the inducing points:
\begin{flalign}
& \ln q(s) = \mathbb{E}_{q\left(\bs f_m\right)\!\!}\left[\ln\mathcal{N}\left(\bs f_m| \bs 0, \bs K_{mm}/s\right)\right] + \ln \mathcal{G}(s; \alpha_0, \beta_0) + \mathrm{const}
= \ln \mathcal{G}(s; \alpha, \beta), & \label{eq:qs}
\end{flalign}
where $\alpha= \alpha_0 + \frac{M}{2}$ and $\beta = \beta_0 + \frac{1}{2}
\textrm{tr}\left(\bs K^{-1}_{mm}\left(S + \hat{\bs f}_m \hat{\bs f}_m^T\right)\right)$.
The expected value is  
$\mathbb{E}[s] = \frac{\alpha}{\beta}$.

We apply variational inference to iteratively reduce the KL-divergence between our approximate posterior
and the true posterior (Equation \ref{eq:svi_approx}) 
by maximising a lower bound, $\mathcal{L}$, on the log marginal likelihood (detailed equations in Appendix \ref{sec:vb_eqns}), which is given by:
\begin{flalign}
&\ln p\left(\bs y | \bs K, \alpha_0, \beta_0\right) = \textrm{KL}\left(q\left(\bs f, \bs f_m, s\right)  || p\left(\bs f, \bs f_m, s | \bs y, \bs K, \alpha_0, \beta_0\right)\right) 
+ \mathcal{L} & \label{eq:lowerbound}
\\
&\mathcal{L} = \mathbb{E}_{q(\bs f)}\left[\ln p(\bs y | \bs f)\right]
+ \mathbb{E}_{q\left(\bs f_m, s\right)}\left[\ln p\left(\bs f_m, s | \bs K, 
\alpha_0, \beta_0 \right) -\ln q\left(\bs f_m\right) - \ln q(s)\right]. & \nonumber
\end{flalign}
To optimise $\mathcal{L}$,
we initialise the $q$ factors randomly, then
update each one in turn, taking expectations with respect to the other factors. 

The only term in $\mathcal{L}$ that refers to the observations, $\bs y$, 
is a sum of $P$ terms, each of which refers to one observation only.
This means that $\mathcal{L}$ can be maximised by considering a random subset of 
observations at each iteration~\citep{hensman2013gaussian}.
For the $i$th update of $q\left(\bs f_m\right)$, we randomly select $P_i$ 
observations $\bs y_i = \{ y_p \forall p \in \bs P_i \}$, 
where $\bs P_i$ is a random subset of indexes of observations,
and $P_i$ is a mini-batch size.
The items referred to by the pairs in the subset are 
$\bs N_i = \{a_p \forall p \in \bs P_i \} \cup \{ b_p \forall p \in \bs P_i\}$.
We  perform updates using $\bs Q_i$ (rows and columns of $\bs Q$ for pairs in $\bs P_i$),
$\bs K_{im}$ and $\bs A_i$ (rows of $\bs K_{nm}$ and $\bs A$ in $\bs N_i$),
$\bs G_i$ (rows of $\bs G$ in $\bs P_i$ and columns in $\bs N_i$), and
$\hat{\bs z}_i = \left\{ \hat{\bs z}_p \forall p \in P_i \right\}$.
The updates optimise the natural parameters of the Gaussian distribution by following the
natural gradient~\citep{hensman2015scalable}:
\begin{flalign}
\bs S^{-1}_i  & = (1 - \rho_i) \bs S^{-1}_{i-1} + \rho_i\left( \mathbb{E}[s]\bs K_{mm}^{-1} + \pi_i\bs A_i^T \bs G^T_{i} \bs Q^{-1}_i \bs G_{i} \bs A_{i} \right)& 
\label{eq:S_stochastic} \\
\hat{\bs f}_{m,i}  & = \bs S_i \left( \! (1 - \rho_i) \bs S^{-1}_{i-1} \hat{\bs f}_{m,i-1}  + 
\rho_i \pi_i  
\bs A_{i}^{T} \bs G_{i}^T \bs Q_i^{-1}\! \left( \bs y_i  - \Phi(\hat{\bs z}_i) + \bs G_{i} \bs A_i \hat{\bs f}_{m,i-1} \! \right) \! \right) & 
\label{eq:fhat_stochastic}
\end{flalign}
where
$\rho_i=(i + \epsilon)^{-r}$ is a mixing coefficient that controls the update rate,
$\pi_i = \frac{P}{P_i}$ weights each update according to sample size,
 $\epsilon$ is a delay hyperparameter and $r$ is a forgetting rate~\citep{hoffman2013stochastic}.

By performing updates in terms of mini-batches, 
the time complexity of Equations \ref{eq:S_stochastic} and
\ref{eq:fhat_stochastic} is
$\mathcal{O}(P_i M^2 + P_i^2 M + M^3)$ and
memory complexity is  $\mathcal{O}(M^2 + P_i^2 + M P_i)$.
The only parameters that must be stored between iterations relate to the 
inducing points, hence the memory consumption does not grow with the dataset size 
as in the GFITC approximation used by \citet{houlsby2012collaborative}.
A further advantage of stochastic updating is that the $s$ parameter (and any other global
parameters not immediately depending on the data) can be learned
before the entire dataset has been processed,
which means that poor initial estimates of $s$ are rapidly improved
and the algorithm can converge faster.

\begin{algorithm}
 \KwIn{ Pairwise labels, $\bs y$, training item features, $\bs x$, 
 test item features $\bs x^*$}
 \nl Select inducing point locations $\bs x_{mm}$ and compute kernel matrices $\bs K$, $\bs K_{mm}$ and $\bs K_{nm}$ given $\bs x$ \;
 \nl Initialise $\mathbb{E}[s]$ and $\hat{\bs f}_m$ to prior means
 and $\bs S$ to prior covariance $\bs K_{mm}$\;
 \While{$\mathcal{L}$ not converged}
 {
 \nl Select random sample, $\bs P_i$, of $P$ observations\;
 \While{$\bs G_i$ not converged}
  {
  \nl Compute $\mathbb{E}[\bs f_i]$ \;
  \nl Compute $\bs G_i$ given $\mathbb{E}[\bs f_i]$ \;
  \nl Compute $\hat{\bs f}_{m,i}$ and $\bs S_{i}$ \;
  }
 \nl Update $q(s)$ and compute $\mathbb{E}[s]$ and $\mathbb{E}[\ln s]$\;
 }
\nl Compute kernel matrices for test items, $\bs K_{**}$ and $\bs K_{*m}$, given $\bs x^*$ \;
\nl Use converged values of $\mathbb{E}[\bs f]$ and $\hat{\bs f}_m$ to estimate
posterior over $\bs f^*$ at test points \;
\KwOut{ Posterior mean of the test values, $\mathbb{E}[\bs f^*]$, and covariance, $\bs C^*$ }
\vspace{0.2cm}
\caption{The SVI algorithm for GPPL: preference learning with a single user.}
\label{al:singleuser}
\end{algorithm}
The complete SVI algorithm is summarised in Algorithm \ref{al:singleuser}.
It uses a nested loop to learn $\bs G_i$, which avoids storing the complete matrix, 
$\bs G$.
It is possible to distribute computation in lines 3-6 by selecting multiple random samples
to process in parallel. A global estimate of $\hat{\bs f}_m$ and $\bs S$
is passed to each compute node, which runs the loop over lines 4 to 6.
The resulting updated $\hat{\bs f}_m$ and $\bs S$ values are then passed back to a 
central node that combines them by taking a mean weighted by $\pi_i$ to account for 
the size of each batch. 

Inducing point locations can be learned
as part of the variational inference procedure, which
breaks convergence guarantees, or by an expensive optimisation process~\citep{hensman2015scalable}. 
We obtain good performance by choosing inducing points up-front 
using K-means++~\citep{arthur2007k} with $M$ clusters to cluster
the feature vectors, 
then taking the cluster centres as inducing points that represent the distribution of observations.

The inferred distribution over the inducing points can be used 
to estimate the posteriors of test items, $f(\bs x^*)$, according to:
\begin{flalign}
\bs f^* \! \! &= \bs K_{*m} \bs K_{mm}^{-1} \hat{\bs f}_m, &
\bs C^* \! \! = \bs K_{**} + \bs K_{*m} \bs K_{mm}^{-1} (\bs S - \bs K_{mm} / \mathbb{E}[s] ) \bs K_{mm}^{-1}\bs K_{*m}^T ,
\end{flalign}
where $\bs C^*$ is the posterior covariance of the test items, $\bs K_{**}$ is their prior covariance, and
$\bs K_{*m}$ is the covariance between test and inducing items.

\subsection{SVI for CrowdGPPL}

We now provide the variational posterior for the crowdGPPL model defined in Equation \ref{eq:joint_crowd}:
\begin{flalign}
& p\left( \bs V, \bs V_m, \bs W, \bs W_m, \bs t, \bs t_m, s^{(v)}_1, .., s^{(v)}_C,
s^{(w)}_1, .., s^{(w)}_C, s^{(t)} | \bs y, \bs X, \bs X_m, \bs U, \bs U_m, k, \alpha_0, \beta_0 \right) 
& \nonumber \\
& \approx q(\bs t) q(\bs t_m)q\left(s^{(t)}\right)\prod_{c=1}^{C} q(\bs v_{c})q(\bs w_c)q(\bs v_{c,m})q(\bs w_{c,m})
q\left(s^{(v)}_c\right)q\left(s^{(w)}_c\right), & 
\end{flalign}
where $\bs U_m$ are the feature vectors of inducing users and the variational $q$ factors are defined below.
We use SVI to optimise the lower bound on the log marginal likelihood 
(detailed in Appendix \ref{sec:crowdL}), which is given by:
\begin{flalign}
& \mathcal{L}_{cr} = 
\mathbb{E}_{q(\bs F)}
[\ln p(\bs y | \bs F)] 
+ \mathbb{E}_{q\left(\bs t_m, s^{(t)}\right)} \left[\ln p\left(\bs t_m, s^{(t)} | \bs K_{mm}, \alpha_0^{(t)}, \beta_0^{(t)}\right)
- \ln q(\bs t_m)  - \ln q\left(s^{(t)}\right) \right]  & \nonumber \\
&
+ \sum_{c=1}^C \!\! \bigg\{  \mathbb{E}_{q\left(\bs v_{m,c},s^{(v)}_c\right)}\left[\ln p\left(\bs v_{m,c}, s^{(v)}_c | \bs K_{mm}, \alpha_0^{(v)}, \beta_0^{(v)}\right) - \ln q(\bs v_{m,c}) - \ln q\left(s_c^{(v)}\right) \right]
&  \nonumber \\ 
& 
+  \mathbb{E}_{q\left(\bs w_{m,c}, s_c^{(w)}\right)}\left[\ln p\left(\bs w_{m,c},s^{(w)}_c | \bs L_{mm}, \alpha_0^{(w)}, \beta_0^{(w)} \right)
  - \ln q(\bs w_{m,c} )  - \ln q\left(s_c^{(w)} \right) \right] \bigg\} . & 
  \label{eq:lowerbound_crowd}
\end{flalign}
The SVI algorithm 
follows the same pattern as Algorithm \ref{al:singleuser}, 
updating each $q$ factor in turn by computing means and covariances
for  $\bs V_m$, $\bs W_m$ and $\bs t_m$ instead of $\bs f_m$ (see Algorithm \ref{al:crowdgppl}).
The time and memory complexity of each update are
$\mathcal{O}(CM_{\mathrm{items}}^3 + CM_{\mathrm{items}}^2 P_i + CM_{\mathrm{items}} P_i^2$
$ + CM_{\mathrm{users}}^3 + CM_{\mathrm{users}}^2 P_i + CM_{\mathrm{users}} P_i^2 )$ 
and 
$\mathcal{O}(CM_{\mathrm{items}}^2 + P_i^2 + M_{\mathrm{items}} P_i + CM_{\mathrm{users}}^2 + M_{\mathrm{users}} P_i)$, respectively.
The variational factor for the $c$th inducing item component is:
\begin{flalign}
\ln q(\bs v_{m,c})  & =  
\mathbb{E}_{q(\bs t, \bs w_{m,c'}\forall c', \bs v_{m,c'}\forall c'\backslash c) }\left[
\ln \mathcal{N}\left( \bs y; \tilde{\Phi}(\bs z), Q \right) \right] 
+ \ln\mathcal{N}\left(\bs v_{\!m,c}; \bs 0, \frac{\bs K_{mm}}{\mathbb{E}\left[s^{(v)}_c\right]}\right) 
 +  \textrm{const} & \nonumber \\
& = \ln \mathcal{N}\left(\bs v_{m,c}; \hat{\bs v}_{m,c}, \bs S_c^{(v)} \right), &
\end{flalign}
where posterior mean $\hat{\bs v}_{m,c}$ and covariance $\bs S_c^{(v)}$ are computed using 
equations of the same form as 
Equations \ref{eq:S_stochastic} and \ref{eq:fhat_stochastic}, except $\bs Q^{-1}$
 is scaled by expectations over $\bs w_{m,c}$,
and $\hat{\bs f}_{m,i}$ is replaced by $\hat{\bs v}_{m,c,i}$.
The factor for the inducing points of $\bs t$ follows a similar pattern to $\bs v_{m,c}$:
\begin{flalign}
\ln q(\bs t_m) & = 
\mathbb{E}_{q(\bs w_{m,c}\forall c, \bs v_{m,c}\forall c)}\left[
\ln \mathcal{N}\left( \bs y; \tilde{\Phi}(\bs z), Q \right) 
\right]
+ \ln\mathcal{N}\left( \bs t_m; \bs 0, \frac{\bs K_{mm}}{\mathbb{E}[s^{(t)}]} \right)
+ \textrm{const} & \nonumber \\
& = \ln \mathcal{N}\left( \bs t_m; \hat{\bs t}_{m}, \bs S^{(t)} \right), & 
\end{flalign}
where the equations for $\hat{\bs t}$ and $\bs S^{(t)}$ 
are the same as Equations \ref{eq:S_stochastic} and \ref{eq:fhat_stochastic}, 
except $\hat{\bs f}_{m,i}$ is replaced by $\hat{\bs t}_{m,i}$. 
Finally, 
the variational distribution for each inducing user's component is:
\begin{flalign}
\ln q(\bs w_{\! m,c} )  = & 
\mathbb{E}_{q(\bs t,\bs w_{m,c'}\forall c'\backslash c, \bs v_{m,c'}\forall c')}\left[
\ln \mathcal{N}\! \left( \bs y; \tilde{\Phi}(\bs z), Q \right) \right] 
+ \ln\mathcal{N}\!\left(\bs w_{\! m,c}; \bs 0, \frac{\bs L_{mm}}{\mathbb{E}[s^{(w)}_c]} \right)
+ \textrm{const} & \nonumber \\
& = \ln \mathcal{N}\left( \bs w_{m,c}; \hat{\bs w}_{\!m,c}, \bs \Sigma_c \right), & 
\end{flalign}
where $\hat{\bs w}_c$ and $\bs \Sigma_{c}$ also follow the pattern of
Equations \ref{eq:S_stochastic} and \ref{eq:fhat_stochastic},
with $\bs Q^{-1}$ scaled by expectations of
$\bs w_{c,m}$,
 and $\hat{\bs f}_{m,i}$ replaced by $\hat{\bs w}_{m,c,i}$.
We provide the complete equations for the variational means 
and covariances for $\bs v_{m,c}$, $\bs t_m$ and $\bs w_{m,c}$ in 
Appendix \ref{sec:post_params}.
The expectations for inverse scales, $s^{(v)}_1,..,s^{(v)}_c$, $s^{(w)}_1,..,s^{(w)}_c$
 and $s^{(t)}$ can be computed using Equation \ref{eq:qs} by
substituting the corresponding terms for $\bs v_c$, $\bs w_c$ or $\bs t$ instead of $\bs f$. 


Predictions for crowdGPPL can be made by computing the posterior mean utilities, $\bs F^*$, 
and the covariance $\bs \Lambda_u^*$ for each user, $u$, in the test set:
\begin{flalign} \label{eq:predict_crowd}
&\bs F^* = \hat{\bs t}^* + \sum_{c=1}^C \hat{\bs v}_{c}^{*T} \hat{\bs w}_{c}^*, \hspace{1cm} \bs \Lambda_u^* = \bs C_{t}^* + \sum_{c=1}^C \omega_{c,u}^* \bs C_{v,c}^* + \hat{w}_{c,u}^2  \bs C_{v,c}^*  +\omega_{c,u}^* \hat{\bs v}_{c}\hat{\bs v}_{c}^T, &
\end{flalign}
where $\hat{\bs t}^*$, $\hat{\bs v}_{c}^*$ and $\hat{\bs w}_{c}^*$ are posterior test means,
$\bs C_{t}^*$ and $\bs C_{v,c}^*$ are posterior covariances of the test items,
and $\omega_{c,u}^*$ is the posterior variance of the user components for $u$. 
(see Appendix \ref{sec:predictions}, Equations \ref{eq:tstar} to \ref{eq:omegastar}).
The mean $\bs F^*$ and covariances $\Lambda^*_u$ can be inserted into Equation \ref{eq:plphi} to predict pairwise labels.
In practice, the full covariance terms are needed only for Equation \ref{eq:plphi}, so need only be computed
between items for which we wish to predict pairwise labels.

\section{Experiments}\label{sec:expts}

\begin{table}[h]
\centering
\small
 \setlength{\tabcolsep}{4pt}
\begin{tabular}{l l l l l l l l l }
\toprule
Dataset & \#folds/ & \#users & total & training set & \multicolumn{2}{c}{test set} & \multicolumn{2}{c}{\#features} \\
              & samples &              & \#items & \#pairs  & \#pairs & \#items
               & items & users \\
\midrule
Simulation a and b & 25 & 25 & 100 & 900 & 0 & 100 & 2 & 2 \\
Simulation c & 25 & 25 & 100 & 36--2304 & 0 & 100 & 2 & 2\\
\midrule
Sushi A-small & 25 & 100 & 10 & 500 & 2500 & 10 & 18 & 123 \\
Sushi A & 25 & 100 & 10 & 2000 & 2500 & 10 & 18 & 123 \\
Sushi B & 25 & 5000 & 100 & 50000 & 5000 & 100 &  18 & 123 \\
\midrule
UKPConvArgCrowdSample & 32 & 1442 & 1052 & 16398 & 529 & 33 & 32310 & 0
\\ \bottomrule
\end{tabular}
\caption{Summary of datasets showing average counts for the training and test sets
used in each fold/subsample. 
The test sets all contain gold-standard rankings over items as well as
pairwise labels, except the simulations, which are not generated as
 we evaluate using the rankings only.
Numbers of features are given after categorical labels have been converted to one-hot encoding, counting
each category as a separate feature.
}
\label{tab:datasets}
\end{table}
Our experiments test key aspects of crowdGPPL: 
 predicting consensus utilities and personal preferences from pairwise labels
 and the scalability of our proposed SVI method.
In Section \ref{sec:exp_synth}, we use simulated data to test the robustness of crowdGPPL
to noise and unknown numbers of latent components.
Section \ref{sec:sushi}
compares different configurations of the model
against alternative methods
using the \emph{Sushi} datasets\footnote{\url{http://www.kamishima.net/sushi/}}~\citep{kamishima2003nantonac}.
Section \ref{sec:exp_scale} evaluates prediction performance and scalability of
 crowdGPPL 
in a high-dimensional
NLP task
with sparse, noisy crowdsourced preferences
(\emph{UKPConvArgCrowdSample}\footnote{\url{https://github.com/ukplab/tacl2018-preference-convincing}}, ~\citet{simpson2018finding}).
Finally, Section \ref{sec:components} evaluates whether crowdGPPL ignores redundant
components.
The datasets are summarised in Table \ref{tab:datasets}.




As baselines, we compare crowdGPPL against 
\emph{GPPL},
which we train on all users' preference labels to learn a single utility function,
and \emph{GPPL-per-user},
in which a separate GPPL instance is learned for each user with no collaborative
learning.
We also compare against the \emph{GPVU} model~\citep{khan2014scalable} 
and 
\emph{collabGP} ~\citep{houlsby2012collaborative}.
CollabGP contains parameters for each pairwise label and
 each user, so has a larger memory footprint than our SVI scheme, 
which stores only the moments at the inducing points.

We test \emph{crowdBT}~\citep{chen2013pairwise} as part of a method for
predicting consensus utilities from crowdsourced pairwise preferences.
CrowdBT models each worker's accuracy, assuming that
the differences between workers' labels are 
due to random errors rather than subjective preferences.
Since crowdBT does not account for the item features,
it cannot predict utilities for items that were not part of the training set.
We therefore treat the posterior mean utilities produced by crowdBT as training labels
for Gaussian process regression using SVI.
We set the observation noise variance of the GP equal to the crowdBT posterior variance
of the utilities 
to propagate uncertainty from crowdBT to the GP. 
This pipeline method, \emph{crowdBT--GP}, 
tests whether it is sufficient to treat annotator differences as noise,
in contrast to the crowdGPPL approach of modelling individual preferences.

We evaluate the methods using the following metrics:
\emph{accuracy (acc)}, which is the fraction of correct pairwise labels;
\emph{cross entropy error (CEE)} between the posterior probabilities over pairwise labels
and the true labels, which captures the quality of the pairwise posterior;
and
\emph{Kendall's $\tau$}, which evaluates the ranking obtained by sorting items by
predicted utility.

\subsection{Simulated Noisy Data}\label{sec:exp_synth}

\begin{figure}[t]
\subfloat[Consensus]{
\label{fig:simB}
\includegraphics[width=.322\columnwidth,clip=true,trim=15 4 13 0]{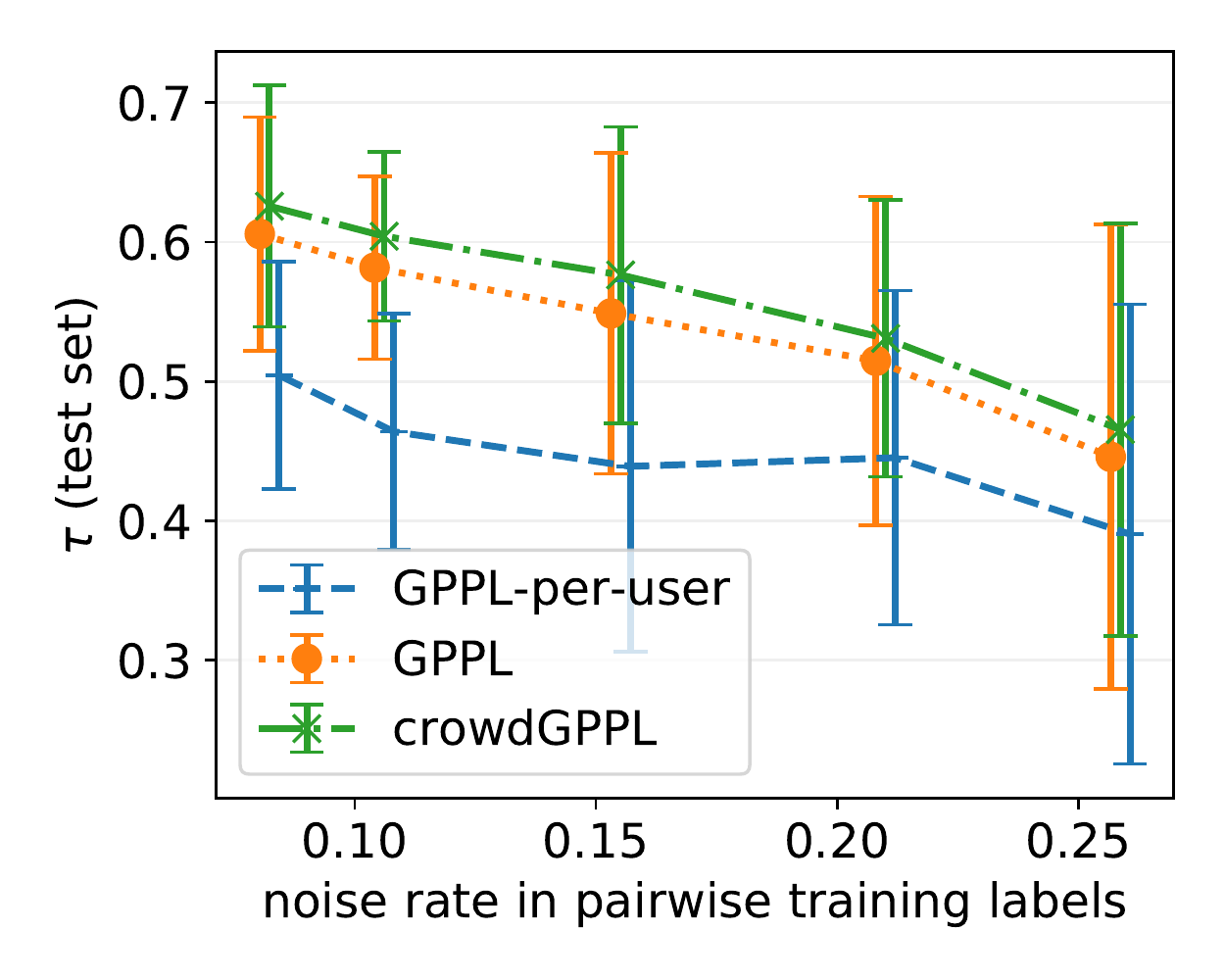}
}
\subfloat[Personal preferences]{
\label{fig:simC}
\includegraphics[width=.306\columnwidth,clip=true,trim=30 5 14 0]{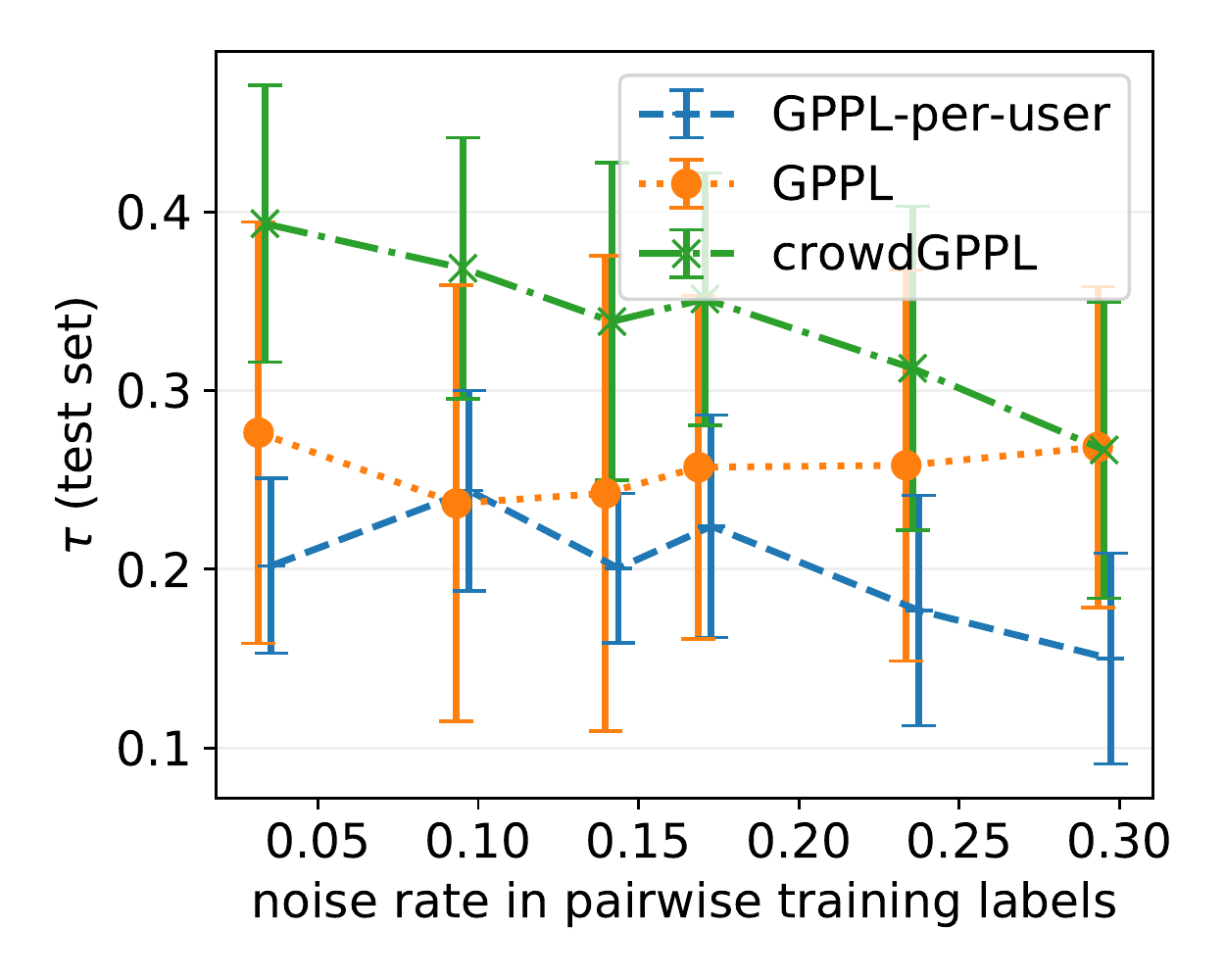}
}
\subfloat[Latent factors]{
\label{fig:simD}
\includegraphics[width=.324\columnwidth,clip=true,trim=8 0 13 0]{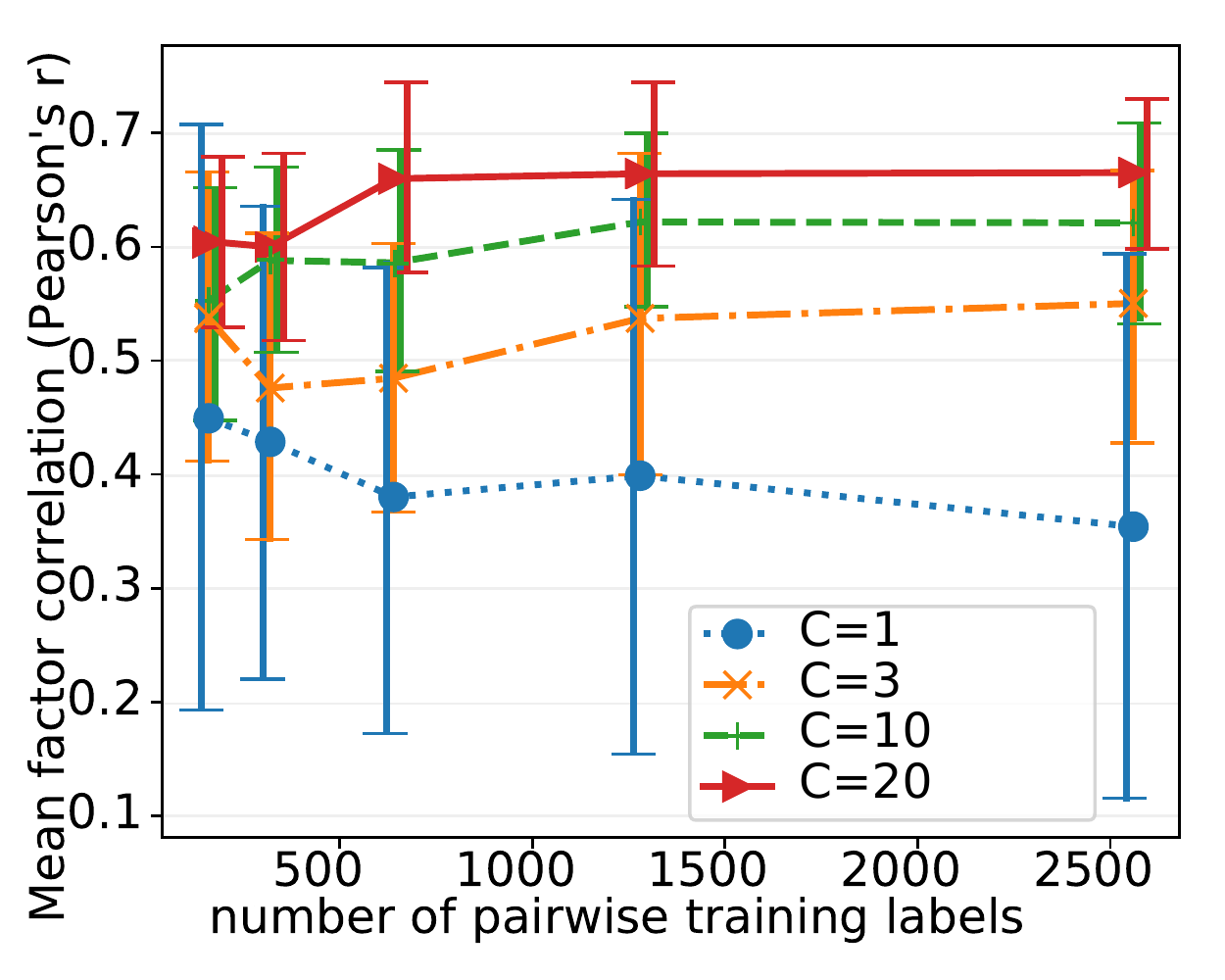}
}
\caption{Simulations: rank correlation between true and inferred utilities.
(a) \& (b) vary the level of noise in pairwise training labels, (c) varies the number of pairwise training labels. 
}
\end{figure}
First, we evaluate whether crowdGPPL is able to model individual preferences
with varying amounts of labelling noise. 
We set the number of latent components to $C=20$ and all Gamma hyperparameters for crowdGPPL, GPPL and GPPL-per-user 
to $\alpha_0 = 1$, $\beta_0 = 100$.
We use Mat\'ern 3/2 kernels with the length-scale for each dimension of the feature vector, $d$,
chosen by a median heuristic:
\begin{flalign}
 l_{d,\mathrm{MH}} = \mathrm{median}( \{ ||x_{i,d} - x_{j,d}||, 
 \forall i=1,..,N, \forall j=1,...,N\} ).
\end{flalign}
This is a computationally frugal way to choose the length-scales,
that has been extensively used in various kernel methods (e.g., ~\citet{bors1996median,gretton2012optimal}).
The SVI hyperparameters were set to 
 $\rho=0.9$, $P_i=1000$ and $\epsilon=1$.
 \citet{hoffman2013stochastic} found that higher values of 
 $\rho$ gave better final results but slightly slower convergence, recommending
 $0.9$ as a good balance across several datasets, 
 and did not find any effect from changing $\epsilon$. 
We follow their recommendations and do not find it necessary to perform further
tuning in our experiments. 
 Both $M$ and $P_i$ are constrained
 in practice by the computational resources available -- we investigate these further in 
 Section \ref{sec:exp_scale}.

In simulation (a), to test consensus prediction,
we generate a $20\times 20$ grid of points
and split them into  50\% training and test sets.
For each gridpoint, we generate pairwise labels by drawing from the generative model of crowdGPPL
with $U=20$ users, $C=5$, each $s^{(v)}_c$ set
to random values between 0.1 and 10, and $s^{(w)}_c = 1, \forall c$.
We vary 
$s^{(t)}$ to control the noise in the consensus function. 
We train and test crowdGPPL with $C=U$ and repeat the complete experiment
$25$ times, including generating new data. 

Figure \ref{fig:simB} shows that crowdGPPL better recovers the 
consensus ranking than the baselines, even as noise increases, as 
GPPL's predictions are worsened by biased users who deviate
consistently from the consensus. 
For GPPL-per-user, the consensus is simply
the mean of all users' predicted utilities, 
so does not benefit from sharing information between users when training.
For simulation (b), we modify the previous setup 
by fixing $s^{(t)} = 5$ and varying $s^{(v)}_c,\forall c$
to evaluate the methods'
ability to recover the personal preferences of simulated users.
The results in Figure \ref{fig:simC} show that crowdGPPL is able to make better 
predictions when noise is below $0.3$.

We hypothesise that crowdGPPL can recover latent components given sufficient training data.
In simulation (c), we generate data using the same setup as before, 
but fix $s^{(t)} = s^{(v)}_c = s^{(w)} = 1,\forall c$
and vary the number of pairwise training labels 
and the number of true components through
$C_{\mathrm{true}} \in \{ 1, 3, 10, 20\}$.
We match inferred components to the true components as follows:
compute Pearson correlations between each unmatched true component and 
each unmatched inferred component;
select the pair with the highest correlation as a match;
repeat until all true components are matched.
In Figure \ref{fig:simD} we plot the mean correlation between matched pairs of components.
For all values of $C_{\mathrm{true}}$, increasing the
number of training labels beyond $700$ brings little improvement. 
Performance is highest when $C_{\mathrm{true}} = 20$,
possibly because the predictive model has $C = 20$,
so is a closer match to the generating model.
However, %
crowdGPPL is able to recover latent components reasonably well for all
values of $C_{\mathrm{true}}$ given $>500$ labels, despite mismatches between $C$ and $C_{\mathrm{true}}$.

\subsection{Sushi Preferences}\label{sec:sushi}

\begin{table}
 \centering
 \small
 \setlength{\tabcolsep}{4pt}
 \begin{tabular}{l l l l@{\hskip 0.5cm} l l l@{\hskip 0.5cm} l l l}
\toprule
& \multicolumn{3}{c}{\textbf{Sushi-A-small}} & \multicolumn{3}{c}{\textbf{Sushi-A}} & \multicolumn{3}{c}{\textbf{Sushi-B}} \\ 
Method & Acc & CEE & $\tau$ & Acc & CEE & $\tau$ & Acc & CEE & $\tau$ \\
\midrule
crowdGPPL & \textbf{.71} & \textbf{.56} & .48 
& .84 & .33 & .79
& .76 & .50 & . 54
 \\
crowdGPPL $\backslash $inducing & .70 & .60 & .45 
& .84 & .34 & .78 
& - & - & - 
\\
crowdGPPL $\backslash  \bs u$ & .70 & .58 & .46 & 
\textbf{.85} & \textbf{.31} & \textbf{.80} 
& \textbf{.78} & .50 & .57 
\\
crowdGPPL $\backslash  \bs u \backslash  \bs x$ & \textbf{.71} & .57 & \textbf{.49} &
\textbf{.85} & .33 & .80 
& .77 & \textbf{.49} & .56 
\\
crowdGPPL $\backslash \bs u,\backslash \bs t$ 
& .68 & .60 & .43 
& .84 & .33 & .80
& .76 & .51 & .58
\\ 
\midrule 
GPPL & .65 & .62 & .31
& .65 & .62 & .31
& .65 & .62 & .31
\\
GPPL-per-user & .67 & .64 & .42
& .83 & .40 & .79 
& .75 & .60 & \textbf{.60} 
\\
collabGP & .69 & .58 & n/a 
& .83 & .35 & n/a
& .76 & \textbf{.49} & n/a
\\
collabGP$\backslash  \bs u$ & .69 & .59 & n/a & .84 & .33 & n/a & .76 & .50 & n/a
\\
GPVU & .70 & .67 & .43 & .72 & .67 & .42 & .73 & .59 & .52
\\ \bottomrule
\end{tabular}
\caption{Predicting personal preferences on \emph{Sushi} datasets,
means over $25$ repeats. 
The standard deviations are $\leq 0.02$ for all accuracies, 
$\leq 0.08$ for all CEE, 
and $\leq 0.03$ for all $\tau$.
For Sushi-B, crowdGPPL, GPPL-per-user and collabGP had runtimes of $~30$ minutes on a 12 core, 2.6GHz CPU server; GPPL required only 1 minute.
 }
\label{tab:sushi}
\end{table}
The sushi datasets contain, for each user, a gold standard preference ranking 
of $10$ types of sushi,
from which we generate gold-standard pairwise labels. 
To test performance with very few training pairs, we obtain \emph{Sushi-A-small}
by selecting $100$ users at random from the complete \emph{Sushi-A} dataset,
then selecting $5$ pairs for training and $25$ for testing per user.
For \emph{Sushi-A}, we select $100$ users at random from the complete dataset, then 
split the data into training and test sets by randomly
selecting $20$ training and $25$ test pairs per user. 
For \emph{Sushi-B}, we use all $5000$ workers, and subsample $10$ training and $1$ test pair per user.

We compare standard crowdGPPL with four other variants: 
\begin{itemize}
\item \emph{crowdGPPL$\backslash$inducing}: does 
not use the sparse inducing point approximation and instead uses all the original 
points in the training set;
\item \emph{crowdGPPL$\mathbf{\backslash \bs u}$}: ignores the user features;
\item  \emph{crowdGPPL$\mathbf{\backslash \bs u \backslash \bs x}$}: ignores both user and item features;
\item  \emph{crowdGPPL$\mathbf{\backslash \bs u \backslash \bs t}$}: 
excludes the consensus function $\bs t$ from the model as well as the user
features. 
\end{itemize}
For methods with $\backslash\bs u$, the user covariance matrix, $\bs L$, 
is replaced by the identity matrix, 
and for crowdGPPL$\mathbf{\backslash \bs u \backslash \bs x}$, 
$\bs K$ is also replaced by the identity matrix.
As the user features do not contain detailed, personal information (only 
region, age group, gender, etc.), they are not expected
to be sufficiently informative to predict personal preferences on their own. 
Therefore, for crowdGPPL and crowdGPPL$\backslash$inducing, 
 we compute $\bs L$ for 10 latent components using
 the Mat\'ern 3/2 kernel function 
 and use the identity matrix for the remaining 10.
CollabGP is also tested with and without user features.
We set hyperparameters $C=20$,
$\epsilon=1$, $\rho=0.9$, $P_i=200$ for  \emph{Sushi-A-small} and \emph{Sushi-A},
 and $P_i=2000$ for \emph{Sushi-B},
without optimisation.
For the gamma hyperparameters,  a grid search over 
$\{10^{-1},...,10^3\}$ on withheld user data from \emph{Sushi-A}
resulted in $\alpha_0=1, \beta_0=100$ for GPPL variants, and 
$\alpha_0^{(t)}=1,\beta_0^{(t)}=100$, 
$\alpha_0^{(v)}=1,\beta_0^{(v)}=10$ and
$\alpha_0^{(w)}=1,\beta_0^{(w)}=10$ for crowdGPPL variants.
The complete process of subsampling, training and testing, was repeated $25$ times
for each dataset.

The results in Table \ref{tab:sushi} 
illustrate the benefit of personalised models over single-user GPPL.
The inducing point approximation does not appear to harm performance of crowdGPPL, but
 including the user features tends to decrease its performance
compared to crowdGPPL$\backslash\bs u$ and crowdGPPL$\backslash\bs u\backslash\bs x$,
except on Sushi-A-small, where they may help with the small amount of training data.
Comparing crowdGPPL$\backslash\bs u$ with crowdGPPL$\backslash\bs u\backslash\bs t$, including the consensus function improves performance modestly.
The strong performance of GPPL-per-user 
suggests that even 10 pairs per person were enough 
to learn a reasonable model for \emph{Sushi-B}.
As expected, the more memory-intensive collabGP performs comparably well to crowdGPPL
on accuracy and CEE but does not provide a ranking function for computing Kendall's $\tau$.
GPVU does not perform as well as other personalised methods on Sushi-A and Sushi-B,
potentially due to its maximum likelihood inference steps.
The results show that crowdGPPL is competitive despite 
the approximate SVI method, 
so in the next experiment, we test the approach on a larger crowdsourced
dataset where
low memory consumption is required.

\subsection{Argument Convincingness}\label{sec:exp_scale}


We evaluate consensus learning, personal preference learning and scalability
on an NLP task, namely, ranking arguments by \emph{convincingness}. 
The task requires learning from crowdsourced data, but is not simply an aggregation task as it 
requires learning a predictor for test documents that were not compared by the crowd.
The dataset, \emph{UKPConvArgCrowdSample}, was subsampled by \citet{simpson2018finding}
from raw data provided by \citet{habernal2016argument}, and
contains arguments written by users
of online debating forums,
with crowdsourced judgements of pairs of arguments
 indicating the most convincing argument.
The data is divided into $32$ folds ($16$ topics, each with 2 opposing stances). For each fold, we train on $31$ folds and test on the remaining fold.
We extend
the task 
to predicting both the consensus and personal preferences of individual crowd workers.
GPPL previously outperformed SVM and Bi-LSTM methods at consensus prediction for \emph{UKPConvArgCrowdSample}~\citep{simpson2018finding}. 
We hypothesise that a worker's view of convincingness 
depends on their personal view of the subject 
discussed, so crowdGPPL may outperform GPPL and
 crowdBT-GP on both consensus and personal preference prediction.

The dataset contains $32,310$ linguistic and embedding features
for each document (we use mean GloVe embeddings for the words in each document, see \citet{simpson2018finding}). The high-dimensionality of the input feature vectors requires us to modify the length-scale heuristic for all GP methods,
as the distance between items grows with the number of dimensions,
which causes the covariance to shrink to very small values. We therefore use
$l_{d,\mathrm{scaledMH}} = 20\sqrt{D} \times l_{d,\mathrm{MH}}$, 
where $D$ is the dimension of the input feature vectors,
and the scale was chosen by comparing the training set accuracy 
with scales in $\{\sqrt{D}, 10\sqrt{D}, 20\sqrt{D}, 100\sqrt{D}\}$.
The hyperparameters are the same as Section \ref{sec:exp_synth} 
except GPPL uses $\alpha_0 = 2$, $\beta_0 = 200$ and
crowdGPPL uses $\alpha^{(t)}_0=\alpha^{(v)}_0=2$, $\beta^{(t)}_0=\beta^{(t)}_0=200$,
$\alpha^{(w)}_0=1$, $\beta^{(w)}_0=10$.
We do not optimise $\alpha_0$, but choose $\beta_0$ by comparing
training set accuracy for GPPL with $\beta_0 \in \left\{2,200,20000\right\}$.
The best value of $\beta_0$ is also used for $\beta^{(t)}_0$ and $\beta^{(v)}_0$, 
then training set accuracy of crowdGPPL is used to select 
$\beta^{(w)}_0 \in \left\{1, 10, 100 \right\}$.
We set $C=50$, $M=500$, $P_i=200$, $\epsilon=10$, and $\rho=0.9$ without optimisation.

\begin{table}
 \centering
 \small
 \setlength{\tabcolsep}{4pt}
\begin{tabular}{ l l l l@{\hskip 1.0cm} l l l@{\hskip 1.0cm} l l l}
\hline
 & \multicolumn{3}{l}{Consensus} & 
 \multicolumn{3}{l}{Personal: all workers} &\multicolumn{3}{l}{$>$50 training pairs} \\
 Method & Acc & CEE & $\tau$ & Acc & CEE & $\tau$ & Acc & CEE & $\tau$ \\ 
  \midrule
 GPPL  & 
 .77 & \textbf{.51} & .50 & 
 .71 &  \textbf{.56} & .31 & 
 .72 &  \textbf{.55} & .25 \\ 
 crowdGPPL & 
 \textbf{.79} & .52 & \textbf{.53} & 
 \textbf{.72} & .58 & \textbf{.33} & 
 \textbf{.74} & \textbf{.55} & \textbf{.27}  \\ 
 crowdGPPL$\backslash \bs t$ & - & - & - &
.68 & .63 & .23 & \textbf{.74} & .57 & \textbf{.27} 
 \\
 crowdBT-GP & .75 & .53 & .45 & .69 & .58 & .30 & .71 & .56 & .23
 \\ \bottomrule
\end{tabular}
\caption{UKPConvArgCrowdSample: predicting consensus, personal preferences for all workers,
and personal preferences for workers with $>$50 pairs in the training set.
}
\label{tab:convarg}
\end{table}
Table \ref{tab:convarg} shows that 
crowdGPPL outperforms both GPPL and
crowdBT--GP 
at predicting both the consensus and personal preferences
(significant for Kendall's $\tau$ with $p<0.05$, Wilcoxon signed-rank test),
suggesting that there is a benefit
to modelling individual workers in subjective, crowdsourced tasks. 
We also compare against crowdGPPL without the consensus (crowdGPPL$\backslash \bs t$)
and find that including $\bs t$ in the model improves personalised
predictions. This is likely because many workers have few training pairs, 
so the consensus helps to identify arguments that are commonly considered very poor or very
convincing.
Table \ref{tab:convarg} also shows that for
workers with more than 50 pairs in the training set,
accuracy and CEE improve for all methods but
$\tau$ decreases,
suggesting that some items may be ranked further away from their correct ranking 
for these workers. It is possible that workers who were willing to complete more
annotations (on average 31 per fold)
deviate further from the consensus, and crowdGPPL does not fully capture
their preferences given the data available.

\begin{figure}[ht]
\centering
\subfloat[Varying $M$]{
\label{fig:M}
 \includegraphics[clip=true,trim=9 0 55 0,width=.28\columnwidth]{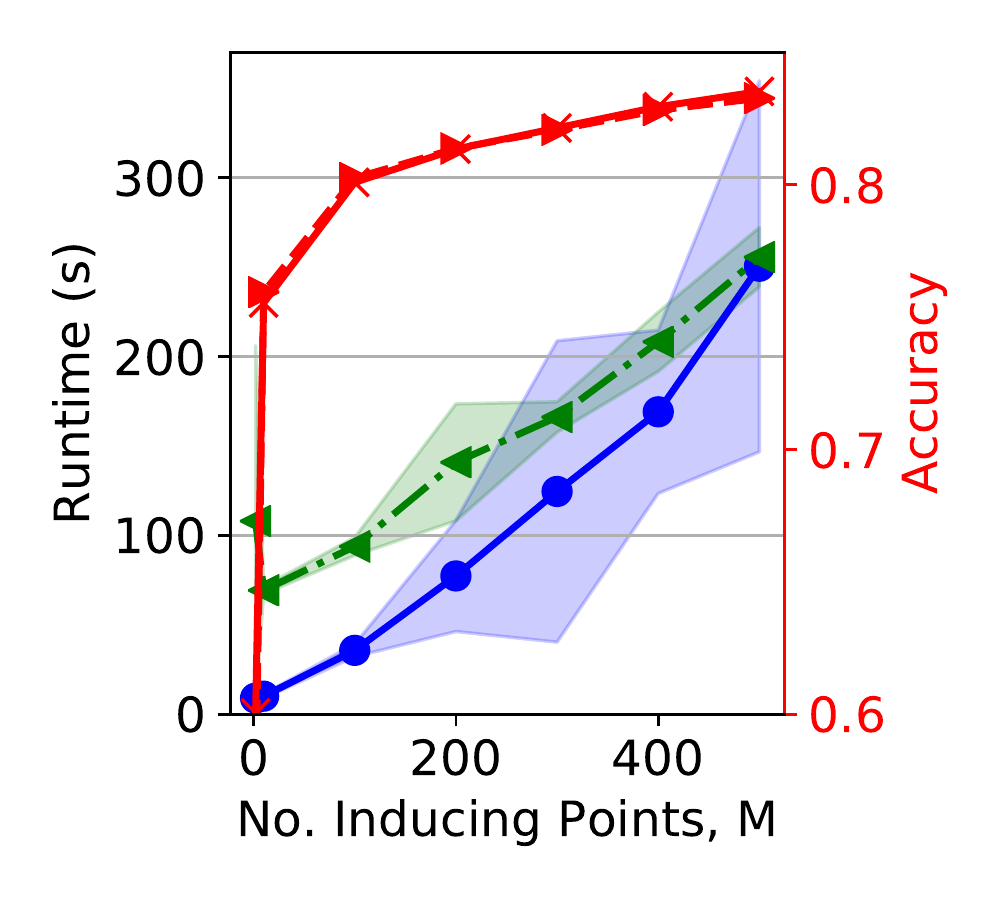}
}
\subfloat[Varying $P_i$   \hspace{0.5cm}]{
\label{fig:P_i}
 \includegraphics[clip=true,trim=60 2 9 0,width=.275\columnwidth]{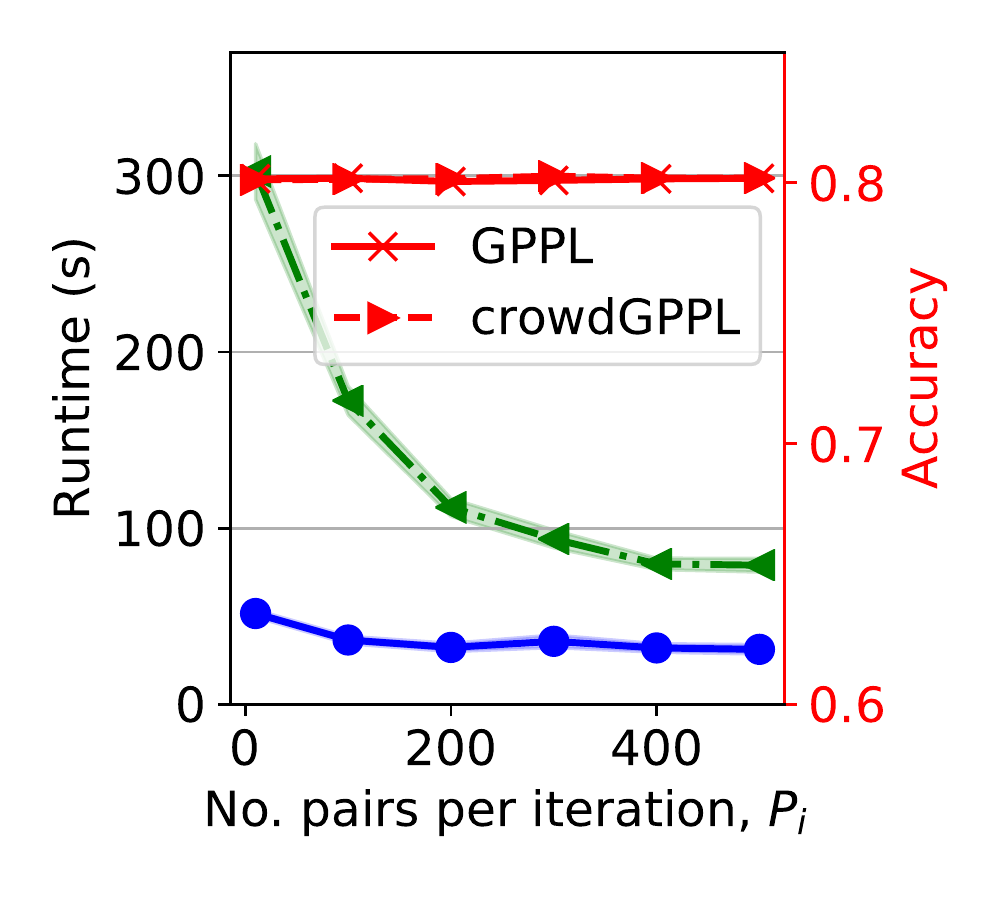}
}\\
\subfloat[Varying $N$]{
\label{fig:Ntr}
\includegraphics[clip=true,trim=9 0 10 0,width=.33\columnwidth]{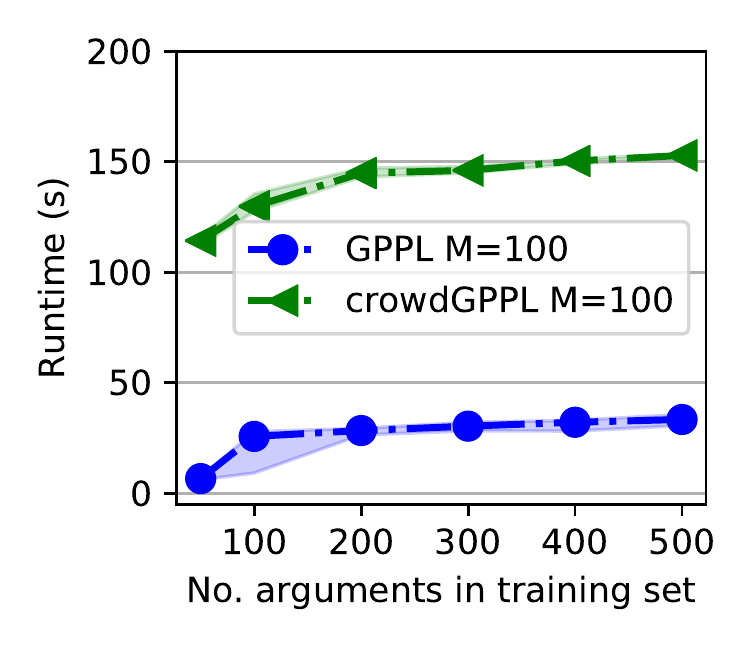}
}
\subfloat[Varying $P$ ]{
\label{fig:Npairs}
\includegraphics[clip=true,trim=26 0 10 0,width=.30\columnwidth]{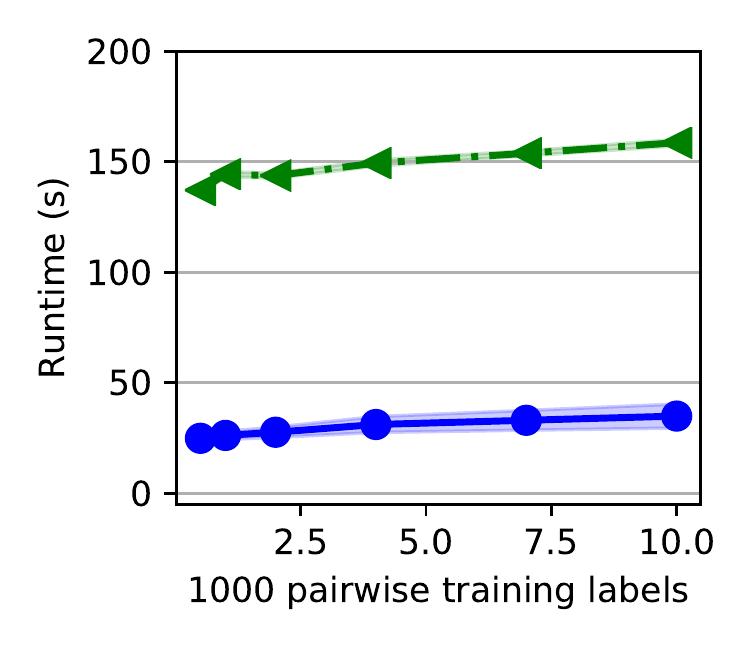}
}
\caption{
    Wall-clock times for training+prediction of consensus utilities for arguments 
    in the training folds of
    UKPConvArgCrowdSample. CrowdGPPL was run with $C=5$. In (b), (c) and (d),  $M=100$.
    Lines show means over 32 runs,
     bands indicate 1 standard deviation (mostly very little variation between folds).
}
\end{figure}
We examine the scalability of our SVI method by evaluating GPPL and crowd-GPPL with
different numbers of inducing points, $M$,
and different mini-batch sizes, $P_i$.
Figure \ref{fig:M} shows the trade-off between
runtime and training set accuracy as an effect of choosing $M$. 
Accuracy levels off as $M$ increases,
while runtime continues to increase rapidly in a polynomial fashion.
Using inducing points can therefore give a large improvement in runtimes 
with a fairly small performance hit.
Figure \ref{fig:P_i} demonstrates that
smaller batch sizes do not negatively affect the accuracy,
although they increase runtimes as more iterations are required for convergence.
The runtimes flatten out as $P_i$ increases, so we recommend choosing $P_i\geq 200$
but small enough to complete an iteration rapidly with the computational resources available.
Figures \ref{fig:Ntr} and \ref{fig:Npairs} show runtimes as a
function of the number of items in the training set, $N$,
and the number of pairwise training labels, $P$, respectively (all other settings remain as in Figure \ref{fig:M}).
In both cases, the increases to runtime are small, despite the growing dataset size.

\subsection{Posterior Variance of Item Components}
\label{sec:components}

\begin{figure}[t]
\centering
\subfloat[\emph{UKPConvArgCrowdSample}]{
\includegraphics[trim=0 5 0 20,clip=true,width=.343\textwidth]{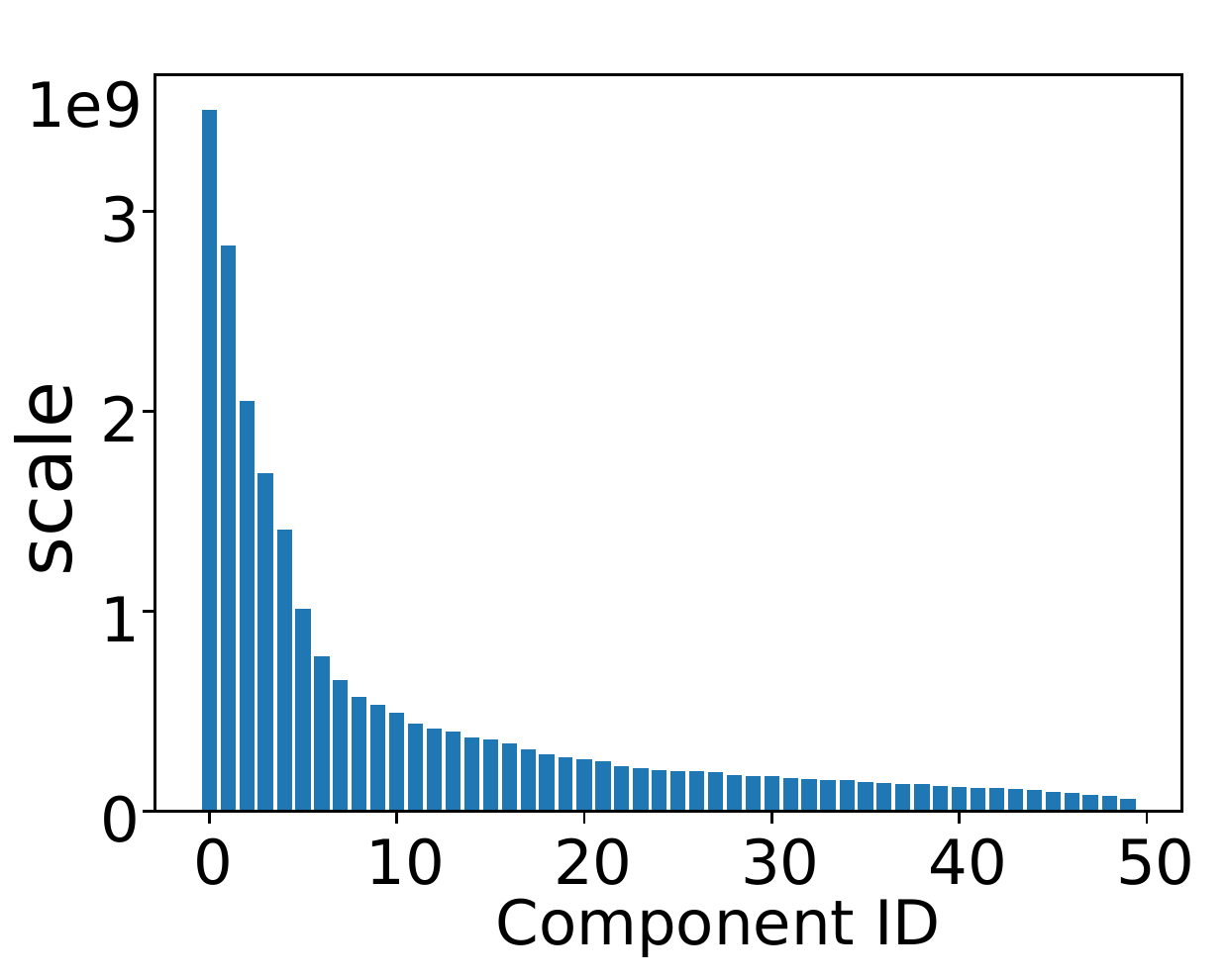}
\hspace{1cm}
}
\subfloat[\emph{Sushi-A}]{
\hspace{1cm}
\includegraphics[trim=10 0 0 17,clip=true,width=.33\textwidth]{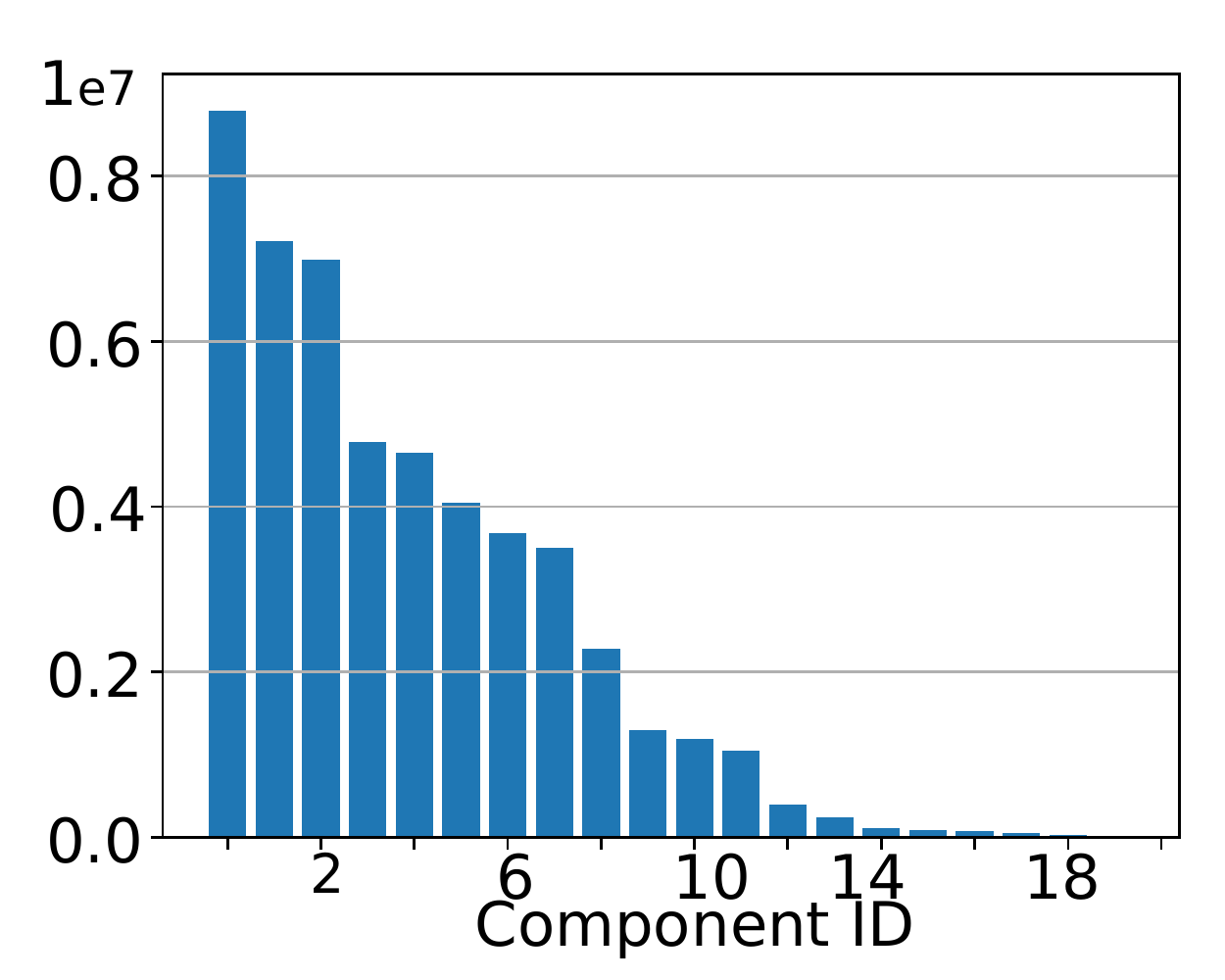}
}
\caption{
Latent component variances, $1/\left(s^{(v)}_c s^{(w)}_c\right)$ in crowdGPPL, means over all runs.
}
\label{fig:latent_factor_variance}
\end{figure}
We investigate how many latent components were actively used by 
 crowdGPPL on the \emph{UKPConvArgCrowdSample} and \emph{Sushi-A} datasets.
Figure \ref{fig:latent_factor_variance}
plots the posterior expectations of the inferred scales, $1/\left(s^{(v)}_c s^{(w)}_c\right)$, for the latent item 
 components. 
 The plots show
that many factors have a relatively small variance and therefore do not contribute 
to many of the model's predictions. This indicates that 
our Bayesian approach will only make use of components that are supported by the data, 
even if $C$ is larger than required.

\section{Conclusions}\label{sec:conclusion}

We proposed a novel Bayesian preference learning approach 
for modelling both the preferences of individuals 
and the overall consensus of a crowd. 
Our model learns the latent utilities of items from pairwise comparisons 
using a combination of Gaussian processes and Bayesian matrix factorisation 
to capture differences in  opinion.
We introduce a stochastic variational inference (SVI) method, that, 
unlike previous work, can scale to arbitrarily large datasets,
since its time and memory complexity do not grow with the dataset size.
Our experiments confirm the method's scalability and
 show that jointly modelling the consensus and personal
preferences can improve predictions of both.
Our approach performs competitively
against less scalable alternatives
and improves on 
the previous state of the art
for predicting argument convincingness from crowdsourced data~\citep{simpson2018finding}.

Future work will investigate learning inducing point locations and
optimising length-scale hyperparameters by maximising 
the variational lower bound, $\mathcal L$,
 as part of the variational inference method.
Another important direction will be to generalise the likelihood from pairwise comparisons
to comparisons involving more than two items~\citep{pan2018stagewise}
or best--worst scaling~\citep{kiritchenko2017best}
to provide scalable Bayesian methods for other forms of comparative preference data.


\section*{Acknowledgements}

This work was supported by the German Federal Ministry of Education and Research (BMBF) 
under promotional references 01UG1416B (CEDIFOR),
by the German Research Foundation through the
the German-Israeli Project Cooperation (DIP, grant DA1600/1-1 and grant GU 798/17-1), 
and by the German Research Foundation EVIDENCE project (grant GU 798/27-1). 
We would like to thank the journal editors and reviewers for their valuable feedback.

\bibliographystyle{spbasic}
\bibliography{simpson_scalable_bayesian_pref_learning_from_crowds}

\appendix

\section{Variational Lower Bound for GPPL}
\label{sec:vb_eqns}

Due to the non-Gaussian likelihood, Equation \ref{eq:plphi},
the posterior distribution over $\bs f$ contains intractable integrals:
\begin{flalign}
p(\bs f | \bs y, k_{\theta}, \alpha_0, \alpha_0) = 
\frac{\int \prod_{p=1}^P \Phi(z_p) \mathcal{N}(\bs f; \bs 0, \bs K_{\theta}/s) 
\mathcal{G}(s; \alpha_0, \beta_0) d s}{\int \int \prod_{p=1}^P \Phi(z_p) \mathcal{N}(\bs f'; \bs 0, \bs K_{\theta}/s) 
\mathcal{G}(s; \alpha_0, \beta_0) d s d f' }.
\label{eq:post_single}
\end{flalign}
We can derive a variational lower bound as follows, beginning with an approximation that does not use inducing points:
\begin{flalign}
\mathcal{L} = \sum_{p=1}^{P} \mathbb{E}_{q(\bs f)}\!\left[ \ln p\left( y_p| f(\bs x_{a_p}), f(\bs x_{b_p}) \right) \right]
\!+ \mathbb{E}_{q(\bs f),q(s)}\!\left[ \ln \frac{p\left( \bs f | \bs 0, \frac{\bs K}{s} \right)}
{q\left(\bs f\right)} \right] 
\!+ \mathbb{E}_{q(s)}\!\left[ \ln \frac{p\left( s | \alpha_0, \beta_0\right)}{q\left(s \right)} \right] &&
\label{eq:vblb}
\end{flalign}
Writing out the expectations in terms of the variational parameters, we get:
\begin{flalign}
\mathcal{L} = &\; \mathbb{E}_{q(\bs f)}\Bigg[ \sum_{p=1}^{P} y_p \ln\Phi(z_p) + (1-y_p) \left(1-\ln\Phi(z_p)\right) \Bigg] 
+ \mathbb{E}_{q(\bs f)}\left[\ln \mathcal{N}\left(\hat{\bs f}; \bs\mu, \bs K/\mathbb{E}[s] \right) \right]
\nonumber\\
& 
- \mathbb{E}_{q(\bs f}\left[\ln\mathcal{N}\left(\bs f; \hat{\bs f}, \bs C \right) \right]
 + \mathbb{E}_{q(s)}\left[ \ln\mathcal{G}\left( s; \alpha_0, \beta_0\right) - \ln\mathcal{G}\left(s; \alpha, \beta \right) \right]  
  \nonumber \\
 =&\;  \sum_{p=1}^{P} y_p \mathbb{E}_{q(\bs f)}[
\ln\Phi(z_p) ]+ (1-y_p) \left(1-\mathbb{E}_{q(\bs f)}[\ln\Phi(z_p)] \right) \Bigg]  \nonumber\\
 & -\frac{1}{2}\left\{
 \ln | \bs K | - \mathbb{E}[\ln s] + \mathrm{tr}\left( \left(\hat{\bs f}^T \hat{\bs f} + \bs C\right)\bs K^{-1} \right)
- \ln |\bs C| - N
 \right\}  \nonumber \\
 & - \Gamma(\alpha_0) + \alpha_0(\ln \beta_0) + (\alpha_0-\alpha)\mathbb{E}[\ln s] + \Gamma(\alpha) + (\beta-\beta_0) \mathbb{E}[s] - \alpha \ln \beta. 
\end{flalign}
The expectation over the likelihood can be computed using numerical integration. 
Now we can introduce the sparse approximation to obtain the bound in Equation \ref{eq:lowerbound}:
\begin{flalign}
\mathcal{L} \approx \; & \mathbb{E}_{q(\bs f)}[\ln p(\bs y | \bs f)]
 + \mathbb{E}_{q(\bs f_m), q(s)}[\ln p(\bs f_m, s | \bs K, 
\alpha_0, \beta_0)] - \mathbb{E}_{q(\bs f_m)}[\ln q(\bs f_m)] 
- \mathbb{E}_{q(s)}[\ln q(s) ] & \nonumber \\ 
=\; & \sum_{p=1}^P \mathbb{E}_{q(\bs f)}[\ln p(y_p | f(\bs x_{a_p}), f(\bs x_{b_p}) )] - \frac{1}{2} \bigg\{ \ln|\bs K_{mm}| - \mathbb{E}[\ln s] - \ln|\bs S| - M
\nonumber &\\
& + \hat{\bs f}_m^T\mathbb{E}[s] \bs K_{mm}^{-1}\hat{\bs f}_m + 
\textrm{tr}(\mathbb{E}[s] \bs K_{mm}^{-1} \bs S) \bigg\}  + \ln\Gamma(\alpha) - \ln\Gamma(\alpha_0)  + \alpha_0(\ln \beta_0) \nonumber\\
& + (\alpha_0-\alpha)\mathbb{E}[\ln s]+ (\beta-\beta_0) \mathbb{E}[s] - \alpha \ln \beta, &
\label{eq:full_L_singleuser}
\end{flalign}
where the terms relating to $\mathbb{E}\left[p(\bs f | \bs f_m) - q(\bs f)\right]$ cancel.
\section{Variational Lower Bound for crowdGPPL}
\label{sec:crowdL}

For crowdGPPL, our approximate variational lower bound is:
\begin{flalign}
\mathcal{L}_{cr} & = \label{eq:lowerbound_crowd_full}
\sum_{p=1}^P \ln p(y_p | \hat{\bs v}_{\!.,a_p}^T \! \hat{\bs w}_{\!.,j_p} \!+ \hat{t}_{a_p}\!,
 \hat{\bs v}_{\!.,b_p}^T\! \hat{\bs w}_{\!.,j_p} \!+ \hat{t}_{b_p})
- \frac{1}{2} 
\Bigg\{  \sum_{c=1}^C \bigg\{  
 \ln|\bs K_{mm}| 
\! - \! \mathbb{E}\left[\ln s^{(v)}_c\right]
\! - \! \ln|\bs S^{(v)}_{c}|  
& \nonumber \\
& 
\! - \! M_{\mathrm{items}} 
+ \hat{\bs v}_{m,c}^T \mathbb{E}\left[s^{(v)}_c\right] \bs K_{mm}^{-1}\hat{\bs v}_{m,c} 
+ \textrm{tr}\left(\mathbb{E}\left[s_c^{(v)}\right] \bs K_{mm}^{-1} \bs S_{v,c}\right) 
+ \ln|\bs L_{mm}|
- \mathbb{E}\left[\ln s^{(w)}_c \right]
& \nonumber \\
&  
- \ln|\bs \Sigma_{c}| 
\! - \! M_{\mathrm{users}}
  + \hat{\bs w}_{m,c}^T \mathbb{E}\left[ s_c^{(w)} \right] \bs L_{mm}^{-1}\hat{\bs w}_{m,c} 
+ \textrm{tr}\left( \mathbb{E}\left[ s_c^{(w)} \right] \bs L_{mm}^{-1} \bs \Sigma_{c} \right)
+ \ln|\bs K_{mm}|   
\bigg\}
& \nonumber \\
&
 - \mathbb{E}\left[\ln s^{(t)} \right]  
- \ln|\bs S^{(t)}| 
- M_{\mathrm{items}} 
+ \hat{\bs t}^T \mathbb{E}\left[s^{(t)}\right] \bs K_{mm}^{-1} \hat{\bs t} 
+ \textrm{tr}\left(\mathbb{E}\left[s^{(t)}\right] \bs K_{mm}^{-1} \bs S^{(t)} \right)
\Bigg\} 
& \nonumber \\
&
+ \sum_{c=1}^C \bigg\{ 
\ln\Gamma\left(\alpha_0^{(v)}\right)  + \alpha_0^{(v)}\left(\ln \beta^{(v)}_0\right)
+ \ln\Gamma\left(\alpha_c^{(v)}\right) + \left(\alpha_0^{(v)} - \alpha_c^{(v)}\right)\mathbb{E}\left[\ln s^{(v)}_c\right]
 & 
\nonumber \\ 
&
+ \left(\beta_c^{(v)} - \beta^{(v)}_0\right) \mathbb{E}[s^{(v)}_c] - \alpha_c^{(v)} \ln \beta_c^{(v)} 
+ \ln\Gamma\left(\alpha_0^{(w)}\right)  + \alpha_0^{(w)}\left(\ln \beta^{(w)}_0\right)
+ \ln\Gamma\left(\alpha_c^{(w)}\right) 
 & 
\nonumber \\ 
&
+ \left(\alpha_0^{(w)} - \alpha_c^{(w)}\right)\mathbb{E}\left[\ln s^{(w)}_c\right]
+ \left(\beta_c^{(w)} - \beta^{(w)}_0\right) \mathbb{E}[s^{(w)}_c] - \alpha_c^{(w)} \ln \beta_c^{(w)} \bigg\}
 + \ln\Gamma\left(\alpha_0^{(t)}\right)  
 & 
\nonumber \\ 
& 
 + \alpha_0^{(t)} \! \left(\ln \beta^{(t)}_0\right)
+  \ln\Gamma\left(\alpha^{(t)}\right) + \left( \! \alpha^{(t)}_0 \!-\! \alpha^{(t)} \! \right)\mathbb{E}\left[\ln s^{(t)}\right]
\! + \!  \left(\! \beta^{(t)} \!-\! \beta^{(t)}_0 \! \right) \mathbb{E}\left[s^{(t)}\right] \! - \!  \alpha^{(t)} \! \ln \beta^{(t)}
. &
\end{flalign}

\section{Posterior Parameters for Variational Factors in CrowdGPPL}
\label{sec:post_params}

For the latent item components, the posterior precision estimate for $\bs S^{-1}_{v,c}$ at iteration $i$ is given by:
\begin{flalign}
\left(  \bs S^{(v)}_{c,i}  \right)^{\!-1} \!\!\! = (1-\rho_i) \left(  \bs S^{(v)}_{c,i-1} \right)^{\!-1} 
\!\! + \rho_i \bs K^{-1}_{mm}\mathbb{E}\left[ s^{(v)}_c \right] \!
+ \rho_i\pi_i \bs A_{i}^T \bs G_i^T \textrm{diag}\left(\hat{\bs w}_{c,\bs u}^2 \! + \bs\Sigma_{c,\bs u,\bs u} \right) 
\bs Q_i^{-1} \bs G_i \bs A_{i}
\!\!, &&
\label{eq:Sv}
\end{flalign}
where $\bs A_{i} = \bs K_{im} \bs K_{mm}^{-1}$, 
$\hat{\bs w}_{c}$ and $\bs\Sigma_{c}$ are the variational mean and covariance of 
the $c$th latent user component (defined below in Equations \ref{eq:what} and \ref{eq:Sigma}),
and ${\bs u} = \{ u_p \forall p \in \bs P_i \}$ is the vector of user indexes in the sample of observations.
We use $\bs S_{v,c}^{-1}$ to compute the means for each row of $\bs V_m$:
\begin{flalign}
\hat{\bs v}_{m,c,i} = & \; \bs S^{(v)}_{c,i}\left( 
(1-\rho_i) \left( \bs S^{(v)}_{c,i-1} \right)^{-1} \hat{\bs v}_{m,c,i-1} \right.& \label{eq:hatv} \\
& \left.
+ \rho_i \pi_i 
\bs S^{(v)}_{c,i} \bs A_{i}^T \bs G_i^T \textrm{diag}(\hat{\bs w}_{c,\bs u}) \bs Q_i^{-1} \right. 
 \left(\bs y_i - \Phi(\hat{\bs z}_i) + \mathrm{diag}(\hat{\bs w}_{c,\bs u}) \bs G_i \bs A_i \hat{\bs v}_{c,m,i-1}^T\right) \bigg). &\nonumber
\end{flalign}

For the consensus, the precision and mean are updated according to the following:
\begin{flalign}
\left( \bs S^{(t)}_i \right)^{-1} = & \; (1-\rho_i) \left( \bs S^{(t)}_{i-1} \right) + \rho_i\bs K^{-1}_{mm}\mathbb{E}\left[s^{(t)}\right] 
+ \rho_i \pi_i \bs A_{i}^T \bs G_i^T \bs Q_i^{-1} \bs G_i \bs A_{i} & \label{eq:St}\\
\hat{\bs t}_{m,i} = & \; \bs S^{(t)}_{i}\left(
(1 - \rho_i) \left( \bs S^{(t)}_{i-1} \right)^{-1}\hat{\bs t}_{m,i-1}  
 + \rho_i \pi_i \bs A_{i}^T \bs G_i^T \bs Q_i^{-1}
\left(\bs y_i - \Phi(\hat{\bs z}_i) + \bs G_i \bs A_{i} \hat{\bs t}_{i} \right) \right). & \label{eq:hatt}
\end{flalign}

For the latent user components, the SVI updates for the parameters are:
\begin{flalign}
\bs \Sigma^{-1}_{c,i} = & \; (1-\rho_i)\bs \Sigma^{-1}_{c,i-1}
+ \rho_i\bs L^{-1}_{mm} \mathbb{E} \left[ s_c^{(w)} \right]
+ \rho_i \pi_i \bs A_{w,i}^T 
& \label{eq:Sigma} \\
& \bigg( \bs H_i^T 
\textrm{diag}\left(\hat{\bs v}_{c,\bs a}^2 
  + \bs S^{(v)}_{c,\bs a, \bs a} + 
\hat{\bs v}_{c,\bs b}^2 + \bs S^{(v)}_{c,\bs b, \bs b}  
- 2\hat{\bs v}_{c,\bs a}\hat{\bs v}_{c,\bs b} - 2\bs S^{(v)}_{c,\bs a, \bs b} \right) \bs Q_i^{-1} 
\bs H_i \bigg) \bs A_{w,i} & \nonumber \\
\hat{\bs w}_{m,c,i} = &\; 
\bs \Sigma_{c,i} \bigg( (1 - \rho_i)\bs \Sigma_{c,i-1}\hat{\bs w}_{m,c,i-1} 
+ \rho_i \pi_i \bs A_{w,i}^T \bs H_i^T \textrm{diag}(\hat{\bs v}_{c,\bs a} - \hat{\bs v}_{c,\bs b})
\bs Q_i^{-1} 
& \label{eq:what} \\
& \Big(\bs y_i - \Phi(\hat{\bs z}_i) + \textrm{diag}(\hat{\bs v}_{c,\bs a} - \hat{\bs v}_{c,\bs b}) \bs H^{(i)}_u \hat{\bs w}_{c,m,i-1}^T\Big) \bigg), & \nonumber
\end{flalign}
where the subscripts $\bs a = \{ a_p \forall p \in P_i \}$
and  $\bs b = \{b_p \forall p \in P_i \}$ are lists of indices to the first and 
second items in the pairs, respectively,
$\bs A_{w,i} = \bs L_{im} \bs L_{mm}^{-1}$,
and $\bs H_i \in U_i \times P_i$ contains partial derivatives of the likelihood corresponding to each user ($U_i$ is the
number of users referred to by pairs in $\bs P_i$), 
with elements given by:
\begin{flalign}
H_{p,j} = \Phi(\mathbb{E}[z_p])(1 - \Phi(\mathbb{E}[z_p])) (2y_p - 1)[j = u_p]. &
\end{flalign}

\begin{algorithm}[h]
 \KwIn{ Pairwise labels, $\bs y$, training item features, $\bs x$, training user features $\bs u$, 
 test item features $\bs x^*$, test user features $\bs u^*$}
 \nl Compute kernel matrices $\bs K$, $\bs K_{mm}$ and $\bs K_{nm}$ given $\bs x$\;
 \nl Compute kernel matrices $\bs L$, $\bs L_{mm}$ and $\bs L_{nm}$ given $\bs u$\;
 \nl Initialise $\mathbb{E} \!\left[s^{(t)}\!\right]$, $\mathbb{E}\!\left[s^{(v)}_c\!\right]\forall c$, 
 $\mathbb{E}\!\left[s^{(w)}_c\!\right]\forall c$, $\mathbb{E}[\bs V]$, $\hat{\bs V}_m$,
 $\mathbb{E}[\bs W]$, $\hat{\bs W}_m$,
  $\mathbb{E}[\bs t]$, $\hat{\bs t}_m$ 
  to prior means\;
 \nl Initialise $\bs S_{v,c}\forall c$ and $\bs S_t$ to prior covariance $\bs K_{mm}$\;
\nl Initialise $\bs S_{w,c}\forall c$ to prior covariance $\bs L_{mm}$\;
 \While{$\mathcal{L}$ not converged}
 {
 \nl Select random sample, $\bs P_i$, of $P$ observations\;
 \While{$\bs G_i$ not converged}
  {
  \nl Compute $\bs G_i$ given $\mathbb{E}[\bs F_i]$ \;
  \nl Compute $\hat{\bs t}_{m,i}$ and $\bs S_{i}^{(t)}$ \;
  \For{c in 1,...,C}
  {
    \nl Update $\mathbb{E}[\bs F_i]$ \;
    \nl Compute $\hat{\bs v}_{m,c,i}$ and $\bs S_{i,c}^{(v)}$ \;
    \nl Update $q\left(s^{(v)}_c\right)$, compute $\mathbb{E}\left[s^{(v)}_c\right]$ and 
    $\mathbb{E}\left[\ln s^{(v)}_c\right]$\; 
    \nl Update $\mathbb{E}[\bs F_i]$ \;
    \nl Compute $\hat{\bs W}_{m,c,i}$ and $\bs \Sigma_{i,c}$ \;    
    \nl Update $q\left(s^{(w)}_c\right)$, compute $\mathbb{E}\left[s^{(w)}_c\right]$ 
    and $\mathbb{E}\left[\ln s^{(w)}_c\right]$\;
  }
  \nl Update $\mathbb{E}[\bs F_i]$ \;
 }
 \nl Update $q\left(s^{(t)}\right)$, compute $\mathbb{E}\left[s^{(t)}\right]$ and
 $\mathbb{E}\left[\ln s^{(t)}\right]$ \;
 }
\nl Compute kernel matrices for test items, $\bs K_{**}$ and $\bs K_{*m}$, given $\bs x^*$ \;
\nl Compute kernel matrices for test users, $\bs L_{**}$ and $\bs L_{*m}$, given $\bs u^*$ \;
\nl Use converged values of $\mathbb{E}[\bs F]$ and $\hat{\bs F}_m$ to estimate
posterior over $\bs F^*$ at test points \;
\KwOut{ Posterior mean of the test values, $\mathbb{E}[\bs F^*]$ and covariance, $\bs C^*$ }
\vspace{0.5cm}
\caption{The SVI algorithm for crowdGPPL.}
\label{al:crowdgppl}
\end{algorithm}

\section{Predictions with CrowdGPPL}
\label{sec:predictions}

The means, item covariances and user variance required for predictions with crowdGPPL (Equation \ref{eq:predict_crowd})
 are defined as follows:
\begin{flalign}
\hat{\bs t}^* & = \bs K_{*m} \bs K^{-1}_{mm} \hat{\bs t}_{m}, \hspace{1.5cm} 
\bs C^{(t)*} = \frac{\bs K_{**}}{\mathbb{E}\left[s^{(t)}\right]} + \bs A_{*m}\left(\bs S^{(t)} \!-\! \bs K_{mm}\right) 
\bs A_{*m}^T, 
\label{eq:tstar} & \\
\hat{\bs v}_{c}^* & = \bs K_{*m} \bs K^{-1}_{mm} \hat{\bs v}_{m,c}, \hspace{1.35cm} 
\bs C^{(v)*}_{c} = \frac{\bs K_{**}}{\mathbb{E}\left[s^{(v)}_c\right ]} + \bs A_{*m} \left(\bs S^{(v)}_{c} 
\!\!-\! \bs K_{mm} \right) \bs A_{*m}^T  & \\
\hat{\bs w}_{c}^* & = \bs L_{*m} \bs L^{-1}_{mm} \hat{\bs w}_{m,c}, \hspace{1.4cm}
\omega_{c,u}^* = 1/ \mathbb{E}\left[s^{(w)}_c \right] + \bs A^{(w)}_{um}(\bs \Sigma_{w,c} - \bs L_{mm}) \bs A^{(w)T}_{um} & \label{eq:omegastar}
\end{flalign}
where  $\bs A_{*m}=\bs K_{*m}\bs K_{mm}^{-1}$,
$\bs A^{(w)}_{um}=\bs L_{um}\bs L_{mm}^{-1}$ and $\bs L_{um}$ is the covariance between user $u$ and the inducing 
users.

\section{Mathematical Notation}
A list of symbols is provided in Tables \ref{tab:sym1} and \ref{tab:sym2}.
\label{sec:not}
\begin{table}[h!]
 \begin{tabularx}{\columnwidth}{p{1.7cm} X }
 \toprule 
 Symbol & Meaning \\
 \midrule 
 \multicolumn{2}{l}{\textbf{General symbols used with multiple variables}} \\
 $\hat{}$ & an expectation over a variable \\
 $\tilde{}$ & an approximation to the variable \\
 upper case, bold letter & a matrix \\
 lower case, bold letter & a vector \\
 lower case, normal letter & a function or scalar \\
 * & indicates that the variable refers to the test set, rather than the training set \\
  \multicolumn{2}{l}{\textbf{Pairwise preference labels}} \\
 $y(a,b)$ & a binary label indicating whether item $a$ is preferred to item $b$ \\
 $y_p$ & the $p$th pairwise label in a set of observations \\
 $\bs y$ & the set of observed values of pairwise labels \\
 $\Phi$ & cumulative density function of the standard Gaussian (normal) distribution \\
 $\bs x_a$ & the features of item a (a numerical vector) \\
 $\bs X$ & the features of all items in the training set \\
 $D$ & the size of the feature vector \\
 $N$ & number of items in the training set \\
 $P$ & number of pairwise labels in the training set \\
 $\bs x^*$ & the features of all items in the test set \\
 $\delta_a$ & observation noise in the utility of item $a$ \\
 $\sigma^2$ & variance of the observation noise in the utilities \\
 $z_p$ & the difference in utilities of items in pair $p$, normalised by its total variance \\
 $\bs z$ & set of $z_p$ values for training pairs \\ 
  \bottomrule
 \end{tabularx}
 \caption{Table of symbols used to represent variables in this paper (continued on next page
 in Table \ref{tab:sym2}).}
 \label{tab:sym1}
\end{table}
\begin{table}
 \begin{tabularx}{\columnwidth}{p{1.7cm} X }
 \toprule 
 Symbol & Meaning \\
 \midrule 
\multicolumn{2}{l}{\textbf{GPPL (some terms also appear in crowdGPPL)}} \\
 $f$ & latent utility function over items in single-user GPPL \\
 $\bs f$ & utilities, i.e., values of the latent utility function for a given set of items \\
 $\bs C$ & posterior covariance in $\bs f$; in crowdGPPL, superscripts indicate 
 whether this is the covariance of consensus values or latent item components \\
 $s$ & an inverse function scale; in crowdGPPL, superscripts indicate which function this variable scales \\
 $k$ & kernel function \\
 $\theta$ & kernel hyperparameters for the items \\
 $\bs K$ & prior covariance matrix over items \\
 $\alpha_0$ & shape hyperparameter of the inverse function scale prior \\
 $\beta_0$ & scale hyperparameters of the inverse function scale prior \\
 \multicolumn{2}{l}{\textbf{CrowdGPPL}} \\
 $\bs F$ & matrix of utilities, where rows correspond to items and columns to users \\
 $\bs t$ & consensus utilities \\
 $C$ & number of latent components \\
 $c$ & index of a component \\
 $\bs V$ & matrix of latent item components, where rows correspond to components \\
 $\bs v_c$ & a row of $\bs V$ for the $c$th component \\
 $\bs W$ & matrix of latent user components, where rows correspond to components \\
 $\bs w_c$ & a row of $\bs W$ for the $c$th component \\ 
 $\bs \omega_c$ & posterior variance for the $c$th user component \\
 $\eta$ & kernel hyperparameters for the users \\
 $\bs L$ & prior covariance matrix over users \\
 $\bs u_j$ & user features for user $j$ \\
 $U$ & number of users in the training set \\
 $\bs U$ & matrix of features for all users in the training set \\ 
 \multicolumn{2}{l}{\textbf{Probability distributions}} \\
 $\mathcal{N}$ & (multivariate) Gaussian or normal distribution \\
 $\mathcal{G}$ & Gamma distribution \\
 \multicolumn{2}{l}{\textbf{Stochastic Variational Inference (SVI) }} \\
 $M$ & number of inducing items \\
 $\bs Q$ & estimated observation noise variance for the approximate posterior \\
 $\gamma, \lambda$ & estimated hyperparameters of a Beta prior distribution over $\Phi(z_p)$ \\
 $i$ & iteration counter for stochastic variational inference \\
 $\bs f_m$ & utilities of inducing items \\
 $\bs K_{mm}$ & prior covariance of the inducing items \\
 $\bs K_{nm}$ & prior covariance between training and inducing items \\
 $\bs S$ & posterior covariance of the inducing items; in crowdGPPL, a superscript and subscript 
 indicate which variable this is the posterior covariance for \\
 $\bs\Sigma$ & posterior covariance over the latent user components \\
 $\bs A$ & $\bs K_{nm} \bs K_{mm}^{-1}$ \\
 $\bs G$ & linearisation term used to approximate the likelihood \\
 $a$ & posterior shape parameter for the Gamma distribution over $s$ \\
 $b$ & posterior scale parameter for the Gamma distribution over $s$ \\
 $\rho_i$ & a mixing coefficient, i.e., a weight given to the $i$th update when combining with current values of variational
 parameters \\
 $\epsilon$ & delay \\
 $r$ & forgetting rate \\
 $\pi_i$ & weight given to the update at the $i$th iteration \\
 $\bs P_i$ & subset of pairwise labels used in the $i$th iteration \\
 $P_i$ & number of pairwise labels in the $i$th iteration subsample \\
 $U_i$ & number of users referred to in the $i$th subsample \\
 $\bs u$ & users in the $i$th subsample \\
 $\bs a$ & indexes of first items in the pairs in the $i$th subsample \\
 $\bs b$ & indexes of first items in the pairs in the $i$th subsample \\
 \bottomrule
 \end{tabularx}
 \caption{Table of symbols used to represent variables in this paper (continued from Table \ref{tab:sym1} on previous page).}
 \label{tab:sym2}
\end{table}


\end{document}